%% file: arxiv.tex
\pdfoutput=1
\documentclass[11pt, letterpaper, logo, onecolumn, numbering]{eorobotics}
\usepackage[comma,authoryear,compress]{natbib}
\definecolor{lightblue}{rgb}{0.22,0.45,0.70}
\hypersetup{
    colorlinks = true,
    citecolor = {lightblue},
    linkcolor = {lightblue},
    urlcolor = {lightblue}
}
\usepackage{subcaption}

\input{preamble}

\makeatletter
\let\ezcam@orig@resizebox\resizebox
\renewcommand{\resizebox}[3]{%
  \sbox\z@{#3}%
  \def\ezcam@argw{#1}%
  \def\ezcam@exclam{!}%
  \ifx\ezcam@argw\ezcam@exclam
    \ezcam@orig@resizebox{#1}{#2}{\usebox\z@}%
  \else
    \ifdim\wd\z@>#1\relax
      \ezcam@orig@resizebox{#1}{#2}{\usebox\z@}%
    \else
      \usebox\z@
    \fi
  \fi
}
\makeatother

\newcommand{\projectpageurl}{https://openfly.art/}
\newcommand{\githuburl}{https://github.com/SHAILAB-IPEC/EZCAM}
\newcommand{\huggingfaceurl}{https://huggingface.co/datasets/IPEC-COMMUNITY/EZCAM}

\definecolor{openflynavy}{HTML}{0A1640}
\definecolor{openflyblue}{HTML}{1F4AD8}
\definecolor{absbg}{HTML}{F4F7FD}
\definecolor{absborder}{HTML}{C2D2EC}
\definecolor{pillbg}{HTML}{F5F8FE}
\definecolor{pillborder}{HTML}{CBD8EE}
\definecolor{affilgray}{HTML}{4A5568}
\definecolor{ruleblue}{HTML}{CBD6E8}

\newcommand{\linkpill}[3]{%
  \href{#2}{%
    \tcbox[on line, boxsep=1.5pt, left=5pt, right=6pt, top=1.5pt, bottom=1.5pt,
           colback=pillbg, colframe=pillborder, boxrule=0.55pt, arc=3.5pt]{%
      #1\hspace{3.5pt}{\sffamily\fontsize{8.5}{10}\selectfont\bfseries\color{openflynavy}#3}%
    }%
  }%
}

\fancypagestyle{firststyle}{
  \fancyhf{}
  \fancyhead[L]{
    \ifthenelse{\boolean{logo}}{
      \includegraphics[height=19pt]{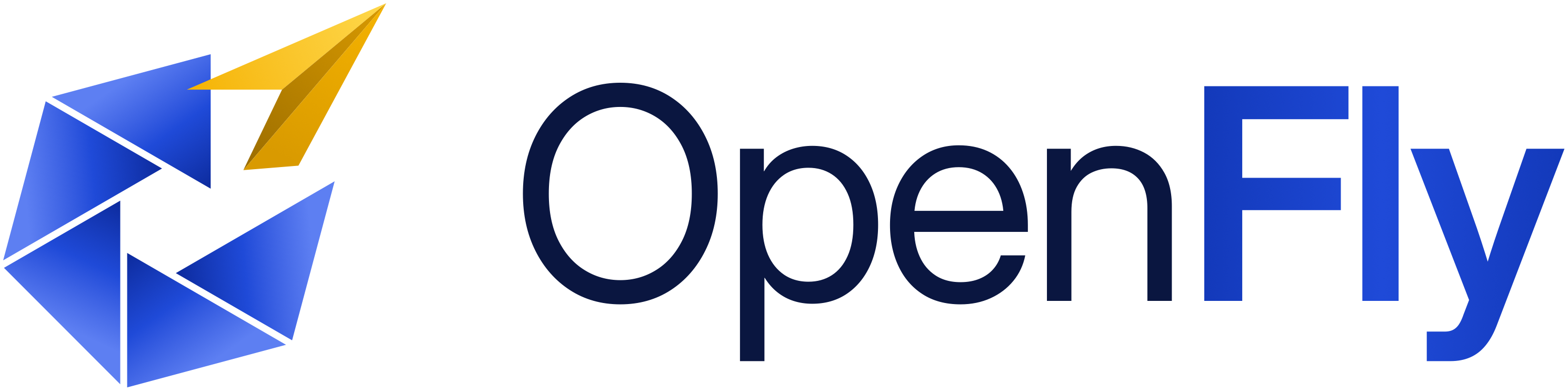}
    }{}
  }
  \fancyhead[R]{
    \ifdefined\paperurl
    \if\relax\the\paperurl\relax \else
      \href{\the\paperurl}{\urlheaderfont \itshape \the\paperurl}\\ \fi
    \else \fi
    {\fontsize{8}{10}\selectfont\itshape\color{affilgray}\monthyeardate\today}
  }
  \renewcommand{\headrulewidth}{0.7pt}
  \renewcommand{\headrule}{\hbox to\headwidth{\color{ruleblue}\leaders\hrule height \headrulewidth\hfill}}
  
}

\def\keye{\raisebox{-1.8pt}{\includegraphics[height=1.02em]{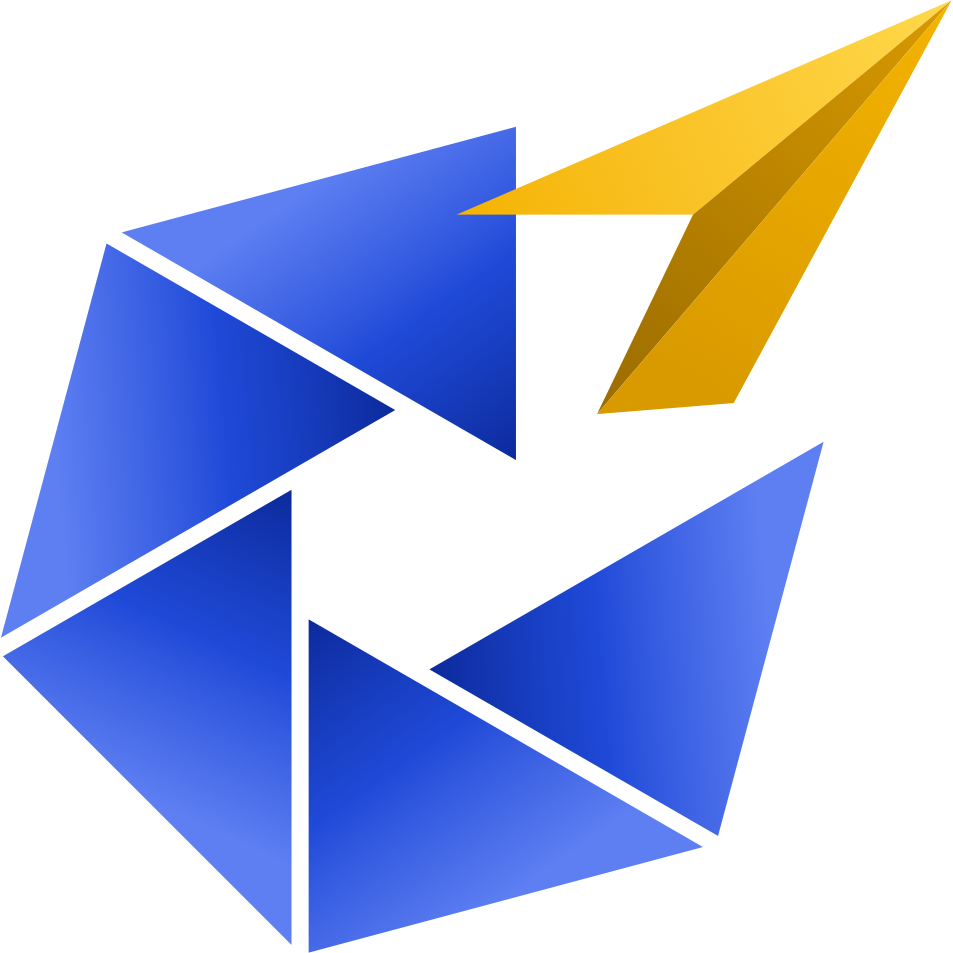}}}
\def\github{\raisebox{-1.8pt}{\includegraphics[height=1.02em]{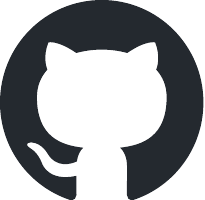}}}
\def\huggingface{\raisebox{-1.8pt}{\includegraphics[height=1.02em]{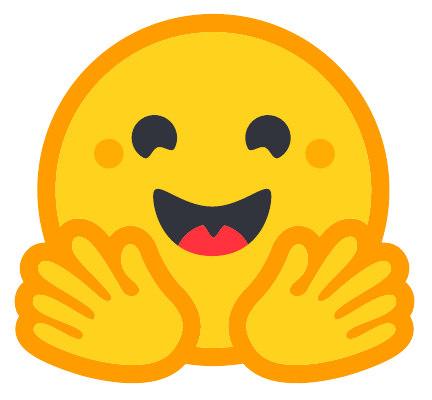}}}

\makeatletter
\renewcommand{\maketitle}{\bgroup\setlength{\parindent}{0pt}\setlength{\parskip}{0pt}%
  {\centering
    \vspace*{-10pt}%
    {\fontsize{18.2}{21.8}\selectfont\bfseries\color{openflynavy}%
      EZCAM: Data, Benchmark, and Policy\\for Embodied Photography\par}%
    \vspace{7pt}%
    {\fontsize{10.0}{12.5}\selectfont\bfseries\color{black}%
      Guanzhou Lan\textsuperscript{1,2,$\ast$}\hspace{0.85em}%
      Xukun Qin\textsuperscript{1,3,$\ast$}\hspace{0.85em}%
      Pingrui Zhang\textsuperscript{1,4}\hspace{0.85em}%
      Yuqi Yang\textsuperscript{5}\hspace{0.85em}%
      Zhigang Wang\textsuperscript{1}\hspace{0.85em}%
      Bin Zhao\textsuperscript{1,2}\par}%
    \vspace{4pt}%
    {\fontsize{8.2}{10.4}\selectfont\color{affilgray}%
      \textsuperscript{1}Shanghai Artificial Intelligence Laboratory\hspace{1.0em}%
      \textsuperscript{2}Northwest Polytechnical University\hspace{1.0em}%
      \textsuperscript{3}Ocean University of China\\[1.5pt]%
      \textsuperscript{4}Fudan University\hspace{1.0em}%
      \textsuperscript{5}China Unidata Logistics Information Co., Ltd.\hspace{1.2em}%
      {\fontsize{7.8}{10.0}\selectfont\textsuperscript{$\ast$}Equal contribution.}\par}%
    \vspace{8pt}%
    \linkpill{\keye}{\projectpageurl}{Project Page}\hspace{8pt}%
    \linkpill{\github}{\githuburl}{GitHub}\hspace{8pt}%
    \linkpill{\huggingface}{\huggingfaceurl}{Hugging Face}\par
  }%
  \vspace{8pt}%
  {\centering
    \captionsetup{type=figure,skip=4pt,font={footnotesize},labelfont={bf}}%
    \includegraphics[width=\textwidth]{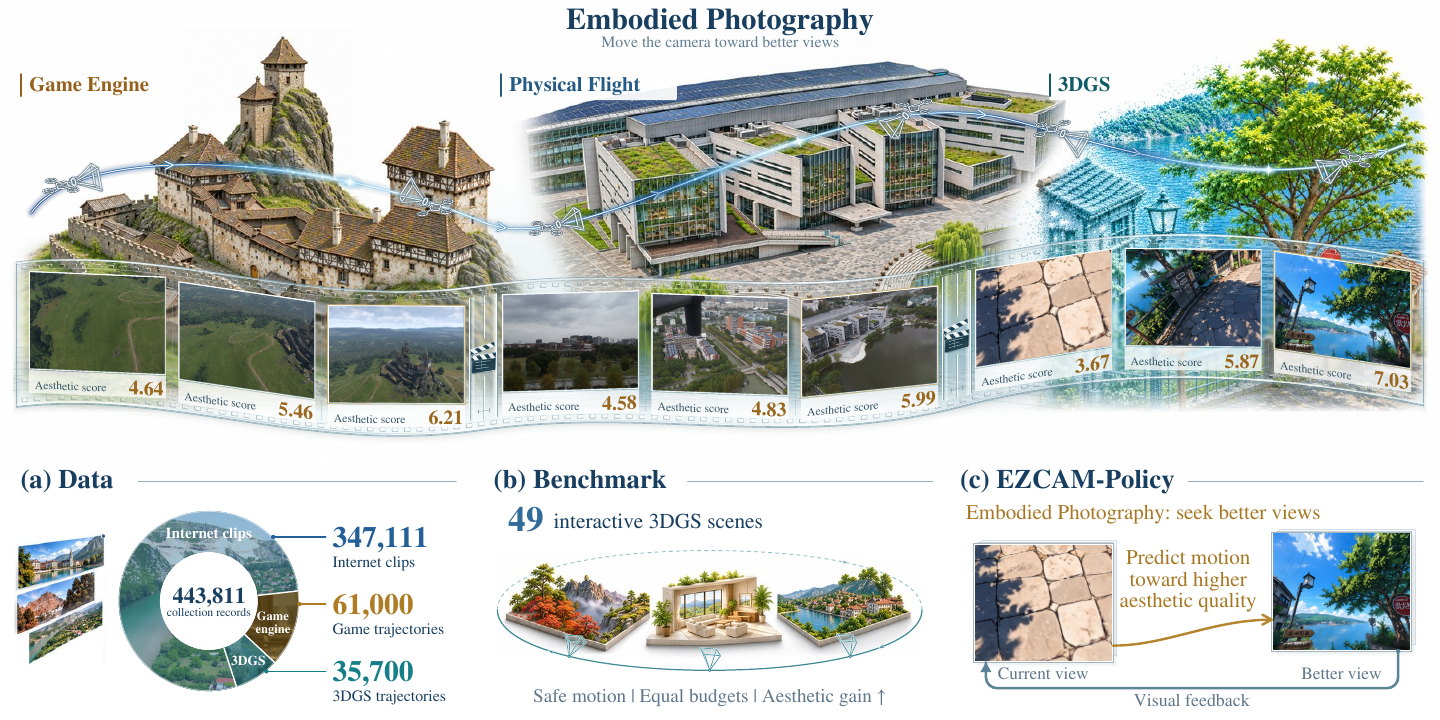}%
    \captionof{figure}{\color{affilgray}\textbf{\color{openflynavy}Overview of EZCAM.} Embodied photography delegates active 6-DoF viewpoint search in 3D space to a closed-loop visual policy:
    \emph{(a)} multi-source score-ascending trajectory curation across Internet videos, game engines, and 3DGS;
    \emph{(b)} standardized 3DGS scene bundles with continuous-path collision checking and unified step budgets; and
    \emph{(c)} \textbf{EZCAM-Policy}, generalizing across 3DGS benchmarks, live game engines, and physical drones.}%
    \label{fig:teaser}\par
  }%
  \vspace{8pt}%
  \begin{tcolorbox}[
    colback=absbg,
    colframe=absbg,
    boxrule=0pt,
    arc=6pt,
    boxsep=1.5pt,
    left=13pt,
    right=13pt,
    top=8pt,
    bottom=6pt,
    before skip=0pt,
    after skip=0pt
  ]
    {\centering\bfseries\color{openflynavy}\fontsize{10.5}{12.5}\selectfont Abstract\par\vspace{4.5pt}}%
    {\linespread{1.05}\fontsize{9.3}{11.1}\color{black!90}\selectfont\theabstract\vspace{12pt}}%
  \end{tcolorbox}
  \egroup
  \thispagestyle{firststyle}
}%
\makeatother

\title{EZCAM: Data, Benchmark, and Policy for Embodied Photography}
\author{}

\begin{document}
\input{sec/0_abstract}
\maketitle
\clearpage

\input{sec/1_intro}
\input{sec/2_related}
\input{sec/2b_formulation}
\input{sec/3_data}
\input{sec/4_benchmark}
\input{arxiv_sec/5_policy}
\input{arxiv_sec/6_experiments}
\input{sec/7_conclusion}

% ===== end sections/conclusion =====
\bibliography{main}

% ===== Inline Supplementary Material (keeps sec/X_suppl.tex unmodified) =====
\clearpage
\makeatletter
\let\ezcam@orig@setcounter\setcounter
\let\ezcam@orig@hypersetup\hypersetup
\def\setcounter#1#2{%
  \def\ezcam@tmpa{#1}\def\ezcam@tmpb{page}%
  \ifx\ezcam@tmpa\ezcam@tmpb\else\ezcam@orig@setcounter{#1}{#2}\fi
}
\def\hypersetup#1{}
\newcommand{\maketitlesupplementary}{%
  \let\setcounter\ezcam@orig@setcounter
  \let\hypersetup\ezcam@orig@hypersetup
  \begin{center}
    {\Large\bfseries Supplementary Material\par}
    \vspace{0.5em}
  \end{center}
}
\makeatother
\input{arxiv_sec/X_suppl}

\end{document}

%% file: preamble.tex
\usepackage{amsfonts}
\usepackage{bm}
\usepackage{float}
\usepackage{stfloats}
\usepackage{placeins}
\usepackage{multirow}
\usepackage{capt-of}
\makeatletter
\def\@dblfloatplacement{\global\@dbltopnum\c@dbltopnumber
   \global\@dbltoproom \dbltopfraction\@colht
   \@textmin \@colht \advance \@textmin -\@dbltoproom
   \@fpmin \dblfloatpagefraction\textheight
   \@fptop \@dblfptop
   \@fpsep \@dblfpsep
   \@fpbot \@dblfpbot
   \def\f@depth{1sp}}%
\makeatother
\usepackage{microtype}

\def\1{\bm{1}}

\DeclareMathAlphabet{\mathsfit}{\encodingdefault}{\sfdefault}{m}{sl}
\SetMathAlphabet{\mathsfit}{bold}{\encodingdefault}{\sfdefault}{bx}{n}

\makeatletter
\renewcommand{\paragraph}{\@startsection{paragraph}{4}{\z@}{5pt plus 1.5pt minus 1pt}{-1em}{\normalsize\bfseries}}
\makeatother

\newcommand{\numtraj}{96{,}700}

\newcommand{\numtotalrecords}{443{,}811}

%% file: sec/0_abstract.tex
\begin{abstract}
Photography represents one of the largest consumer markets worldwide. However, capturing compelling photographs has long relied on human judgment to actively adjust 3D viewpoints and framing, giving rise to professional photographers.
We introduce \textbf{Embodied Photography}, which delegates this decision-making process to AI: an agent observes ego-centric RGB views and executes continuous 6-DoF actions in 3D space to seek stronger compositions.
Lacking interactive testbeds and aligned trajectory data, prior work relies on 2D edits, open-loop generation, or per-scene search rather than generalizable 3D policies.
We build \textbf{EZCAM}:
\textbf{(1) Data:} Automated curation and graph search across Internet videos, game engines, and 3D Gaussian Splatting (3DGS) yield \textbf{443,811} score-ascending records (347,111 Internet clips, 61,000 game, and 35,700 3DGS trajectories) pairing RGB streams with 6-DoF motions.
\textbf{(2) Benchmark:} A 49-scene 3DGS interactive benchmark enables free 6-DoF capture under continuous-path 3D collision checking and unified step budgets.
\textbf{(3) Policy and Insights:} We train \textbf{EZCAM-Policy}, a flow-matching Diffusion Transformer with an aesthetic-pretrained encoder and score conditioning that predicts 32-step 6-DoF action chunks for closed-loop control with dynamic onboard top-$K$ shutter selection.
Its design reflects two insights: \emph{(i)} intermediate transitions have locally non-monotonic aesthetics, making long-horizon chunk prediction essential; and \emph{(ii)} aesthetic pretraining and score features supply compositional priors absent in navigation backbones.
\textbf{EZCAM-Policy} outperforms baselines across unseen aesthetic scorers on our benchmark, wins \textbf{93.8\%} of 1,590 blinded human preferences ($N=106$) over per-scene best baselines, and transfers zero-shot to live game engines and a DJI Matrice 30T drone with NVIDIA Orin NX.\par\vspace{-12pt}
\end{abstract}

%% file: sec/1_intro.tex
\section{Introduction}
\label{sec:intro}
From professional cinematography to smartphones and aerial drones, photography constitutes a massive global market.
Even in the AIGC era, people retain an irreplaceable demand for visual authenticity to record genuine physical moments, yet capturing compelling real-world photographs still relies on human expertise to adjust 6-DoF camera viewpoints in 3D space.
We introduce \textbf{Embodied Photography} to delegate this active framing process to AI: from an initial view in a 3D scene, a policy iteratively observes recent RGB frames and executes continuous 6-DoF camera movements to seek stronger compositions within a finite interaction budget.

This shifts the paradigm from \emph{computational photography} to \emph{embodied photography}, evolving \textbf{passive image enhancement and IQA} into \textbf{active viewpoint selection and criterion execution.}
Progress toward this goal has been bottlenecked by the lack of photorealistic interactive testbeds and large-scale aligned camera trajectories, leading prior work to rely on three shortcuts:
\emph{(i) 2D or single-view adjustments}~\citep{wei2018goodview,zeng2019gaic,li2025spas,yao2025ppc,you2026photoframer}, which crop sub-windows, predict local rotational nudges, or synthesize views from one image without 3D motion;
\emph{(ii) discrete-grid RL or test-time search}~\citep{alzayer2021autophoto,tang2026aestheticfield,che2026photoagent,jiang2026beforeshutter,guo2026photoflow}, which rely on online trial-and-error in simplified simulators~\citep{savva2019habitat} or per-scene search over pre-scanned 3D fields rather than learning a generalizable continuous 6-DoF policy; and
\emph{(iii) offline-evaluated trajectory generation}~\citep{li2024director3d,zhang2025gendop,ren2025gen3c,xing2025motioncanvas,courant2024etdirector,liu2024chatcam,jiang2024ccd}, which evaluates generated paths via open-loop trajectory error or video statistics (e.g., FVD) rather than interactive 3D framing quality.

We establish embodied photography as a data-driven, closed-loop sequential decision-making paradigm through \textbf{EZCAM} (Figure~\ref{fig:teaser}).
To supply compositional trajectory supervision absent in standard navigation logs, we build multi-source pipelines that construct 6-DoF paths progressing toward higher aesthetics; to evaluate closed-loop 3D framing at minimal cost, we package calibrated 3D Gaussian Splatting (3DGS) scenes into collision-checked, budget-accounted testbeds; and to learn closed-loop control, we design a flow-matching Diffusion Transformer grounded in two insights: \emph{(i)} intermediate transitions are locally non-monotonic, making long-horizon chunk prediction essential, and \emph{(ii)} aesthetic pretraining provides compositional priors absent in navigation backbones.

\paragraph{Contributions.}\leavevmode
\begin{enumerate}[label=(\arabic*), leftmargin=*, itemsep=2pt, topsep=3pt]
    \item \textbf{Data:} Automated curation and graph-search pipelines across Internet videos, game engines, and 3DGS, yielding \textbf{\numtotalrecords{}} score-ascending records (347,111 Internet clips, 61,000 game trajectories, and 35,700 3DGS trajectories).
    \item \textbf{Benchmark:} A 49-scene 3DGS interactive benchmark with continuous-path 3D collision checking, unified step budgets, and a standardized scene interface.
    \item \textbf{Policy, Insights, and Deployment:} \textbf{EZCAM-Policy}, built on long-horizon chunk prediction and aesthetic pretraining, winning \textbf{93.8\%} of blinded human judgments ($N=106$) over per-scene best baselines and transferring zero-shot to live game engines and DJI Matrice 30T aerial drones.
\end{enumerate}

\begin{figure*}[t]
\centering
\includegraphics[width=0.98\textwidth]{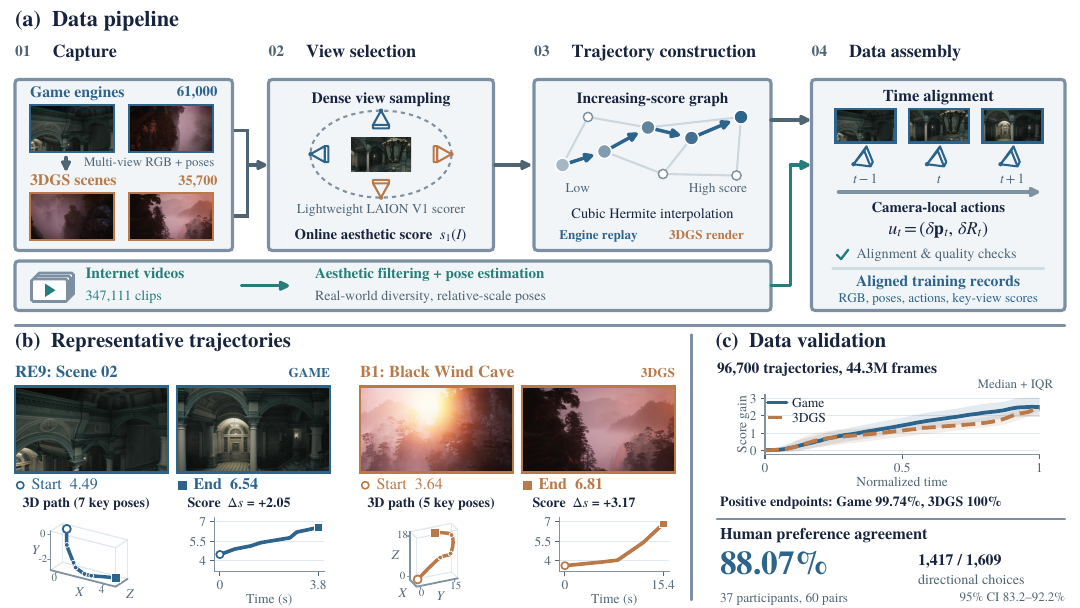}\vspace{-5pt}
\caption{\textbf{Multi-source data construction, 3D trajectory structure, and validation.}
\emph{(a)} Automated pipelines across game engines, 3DGS, and Internet videos score views via $s_1(I)$ and align ascending camera paths.
\emph{(b)} Representative game (\emph{Resident Evil Requiem}, RE9) and 3DGS 3D trajectories (circle: start; square: end).
\emph{(c)} Full-frame cross-scorer progression and blinded human preference alignment.}
\label{fig:data-human-validation}\label{fig:data-keyview-examples}\label{fig:data-full-corpus}
\vspace{-7pt}
\end{figure*}

%% file: sec/2_related.tex
\section{Related Work}
\label{sec:related}
\paragraph{Aesthetics and composition.}
Image-rating datasets and quality predictors~\citep{murray2012ava,kong2016aadb,talebi2018nima,radford2021clip} supervise still-photo scoring on fixed captures, while cropping and composition-guidance methods~\citep{wei2018goodview,zeng2019gaic,li2025spas,yao2025ppc,you2026photoframer} adjust 2D sub-windows or suggest local rotational nudges from an existing view.
Because post-hoc 2D edits and local nudges cannot alter 3D perspective, foreground-background parallax, or physical occlusions, EZCAM studies closed-loop 6-DoF actions that actively relocate the camera in 3D space, while separating fast online collection scoring from held-out multi-model and human evaluation.

\paragraph{Camera control and embodied policies.}
Generative camera-trajectory models~\citep{zhang2025gendop,li2024director3d,ren2025gen3c,xing2025motioncanvas,courant2024etdirector,liu2024chatcam,jiang2024ccd} and scene-tour datasets~\citep{zhou2018stereo,wang2025spatialvid} capture cinematic motion priors, yet synthesize paths open-loop and evaluate against reference poses or video statistics without interactive feedback, while aerial cinematography~\citep{bonatti2020autonomous} tracks designated targets over local maps.
Embodied navigation and VLA policies~\citep{anderson2018r2r,zhang2024navid,zhang2025uninavid,brohan2023rt2,kim2024openvla}, alongside visuomotor chunking and diffusion/flow controllers~\citep{zhao2023act,chi2023diffusionpolicy,black2024pi0}, map observations to actions for semantic goal reaching or manipulation; embodied photography instead shifts closed-loop chunked control to map-free aesthetic viewpoint discovery from non-monotonic ascending trajectories.
Active vision~\citep{bajcsy1988active,connolly1985nbv,zeng2020nbvsurvey,sakamoto2026e3vs} similarly couples sensing and motion for geometric coverage or VQA, whereas recent 3D photography planners~\citep{alzayer2021autophoto,tang2026aestheticfield,che2026photoagent,jiang2026beforeshutter,guo2026photoflow,wu2026robophotovla} require discrete grids, pre-scanned 3D fields, scene graphs, or uncharged previews~\citep{tang2026aestheticfield,jiang2026beforeshutter,guo2026photoflow} rather than zero-map RGB control under a shared observation budget.

\paragraph{View synthesis as an interaction environment.}
Neural rendering and 3DGS~\citep{mildenhall2020nerf,kerbl20233dgs} make captured scenes~\citep{barron2022mipnerf360,knapitsch2017tanks,hedman2018deepblending} photorealistically queryable at novel 6-DoF poses alongside mesh simulators~\citep{savva2019habitat}.
Because unconstrained radiance fields permit invalid queries inside obstacles or under-reconstructed regions that corrupt visual scoring, EZCAM pairs standardized 3DGS bundles with continuous-path collision checking and preview-free budget accounting, decoupling geometric validity, observation cost, and aesthetics.

%% file: sec/2b_formulation.tex
\section{Problem Formulation}
\label{sec:task}
Given a 3D world $\mathcal{W}$, we formulate \emph{embodied photography} as a map-free, closed-loop sequential decision-making task that maximizes the aesthetic score $s(I)$ of captured photographs over an interaction budget $T$:
\begin{equation}
 \begin{aligned}
  \max_{\pi_\theta}\; \mathbb{E}\!\left[\max_{0\leq t\leq T} s(I_t)\right] \quad
  \text{s.t.}\;\; I_{t+1} &= \mathcal{R}\!\big(\mathcal{W},\, u_{0:t}\big), \\[-2pt]
  u_{t:t+k-1} &= \pi_\theta\big(I_{t-h+1:t}, u_{t-h:t-1}\big).
 \end{aligned}
 \label{eq:intro-embodied-photo}
\end{equation}
Starting from an initial observation $I_0$, at each step $t$ the policy $\pi_\theta$ conditions on the past $h$ RGB frames $I_{t-h+1:t}$ and executed camera-local actions $u_{t-h:t-1}$ to predict $k$ relative 6-DoF actions $u_{t:t+k-1}$ in $\mathrm{SE}(3)$, without underlying 3D geometry, global poses, or uncharged future previews.
Executing $u_t = (\delta\mathbf{p}_t, \delta R_t) \in \mathrm{SE}(3)$ updates the camera viewpoint within the collision-free space of $\mathcal{W}$ and renders $I_{t+1} = \mathcal{R}(\mathcal{W}, u_{0:t})$, consuming one unit of the budget $T$.

%% file: sec/3_data.tex
\section{Data}
\label{sec:data}
This section presents our data foundation for Eq.~\ref{eq:intro-embodied-photo}.
We introduce our aesthetic scorers (Section~\ref{sec:data-scorers}) and data pipelines (Section~\ref{sec:data-pipelines}), verify cross-scorer progression (Section~\ref{sec:data-validation-main}), and summarize corpus properties (Section~\ref{sec:data-characteristics}).

\subsection{Aesthetic Scoring and Human Alignment}
\label{sec:data-scorers}\label{sec:data-human}
Photographic aesthetics use $[0,10]$-aligned AIGC scorers: the fast two-head LAION V1 predictor $s_1(I)$~\citep{schuhmann2022laionaesthetics}, SigLIP V2.5 $s_{2.5}(I)$~\citep{discus0434aestheticv25} (clipped to $[0,10]$), and MLLM ArtiMuse $s_{\mathrm{AM}}(I)$~\citep{cao2025artimuse} (scaled by $1/10$ to $[0,10]$).
Collection, filtering ($\Delta s_1>0$), and fast online runs use $s_1$; $s_{2.5}$ and $s_{\mathrm{AM}}$ remain unseen during collection and policy training.
Evaluation uses the equal $1/3$ ensemble $s_{\mathrm{ens}}(I)=\frac{1}{3}[s_1(I)+s_{2.5}(I)+s_{\mathrm{AM}}(I)]$ and the strictly LAION-free $s_{\mathrm{indep}}(I)=\frac{1}{2}[s_{2.5}(I)+s_{\mathrm{AM}}(I)]$ (Appendix~\ref{app:data-models}).
Blinded human judgments ($N=37$) match $s_{\mathrm{ens}}$ rankings in \textbf{88.07\%} of directional choices (Figure~\ref{fig:data-human-validation}c; Appendix~\ref{app:data-human}), confirming human preference alignment.

\begin{figure*}[t]
\centering
\includegraphics[width=0.98\textwidth]{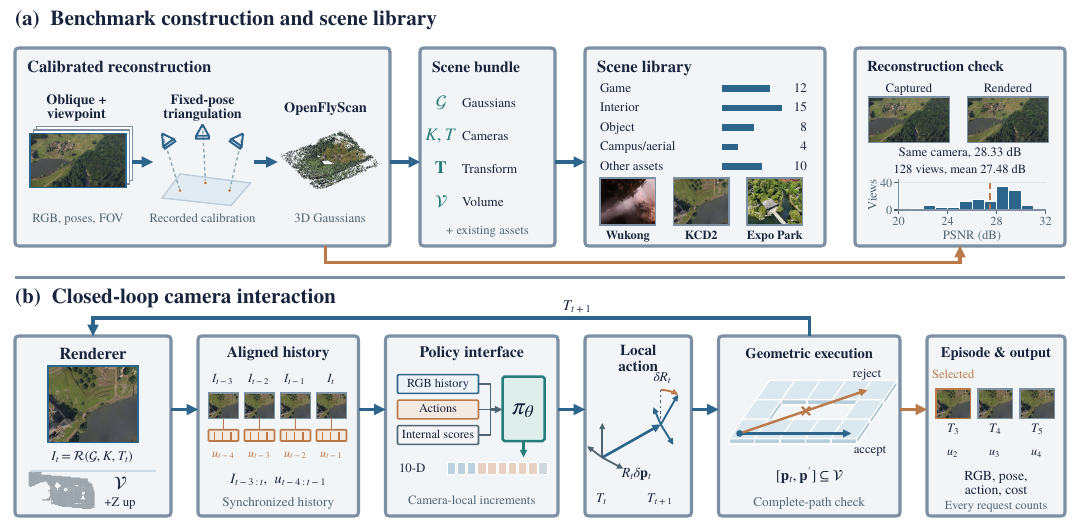}\vspace{-5pt}
\caption{\textbf{Benchmark construction and closed-loop evaluation infrastructure.}
\emph{(a)} Calibrated multi-view 3DGS reconstruction and standardized plug-and-play scene bundles.
\emph{(b)} Closed-loop EWA rendering with continuous-segment collision verification and strict budget accounting.}
\label{fig:data-infra}\label{fig:infrastructure}
\vspace{-7pt}
\end{figure*}

\subsection{Multi-Source Data Pipelines}
\label{sec:data-pipelines}
Training a generalizable policy requires broad visual diversity and metric 6-DoF supervision, which no single source offers alone. We build three pipelines (Figure~\ref{fig:data-keyview-examples}a) totaling \textbf{\numtotalrecords{}} records (Table~\ref{tab:data-main-sources}, Appendix~\ref{app:data-records}):
\emph{(1) Internet videos (347,111 clips):} We sample SpatialVID~\citep{wang2025spatialvid} frames via $s_1(I)$, locate smooth ascending segments, and crop uniform clips with relative-scale actions $u_t = (\delta\mathbf{p}_t, \delta R_t)$.
\emph{(2) Game engines (61,000 trajectories):} Our general UE5 capture platform scans collision-free views across Unreal titles, scores each via $s_1(I)$, searches directed ascending graphs ($\Delta s_1 > 0$), and applies cubic Hermite interpolation to render continuous metric 6-DoF paths.
\emph{(3) 3DGS scenes (35,700 trajectories):} Bridging to radiance-field testbeds, we apply the same $s_1(I)$ graph search and Hermite replay in reconstructed 3DGS scenes (Figure~\ref{fig:data-keyview-examples}b; Appendix~\ref{app:data-release}).
Because interpolated transit frames between ascending key views need not strictly rise and per-frame monotonicity is combinatorially intractable, we enforce an \textbf{overall upward trend} and score all decoded frames afterward.

\subsection{Verifying Cross-Scorer Aesthetic Progression}
\label{sec:data-validation-main}
During data collection, we use only lightweight $s_1(I)$ to ensure high speed and efficiency across millions of candidate views.
After trajectory collection is completed, to rigorously verify the validity of the constructed data, we score every decoded frame (\textbf{44,264,749} frames across all \numtraj{} trajectories) with all three aesthetic models ($s_1$, $s_{2.5}$, and $s_{\mathrm{AM}}$), confirming consistent progression under both unseen scorers and $s_{\mathrm{ens}}(I)$ (Figure~\ref{fig:data-full-corpus}c; Appendix~\ref{app:data-audit}, Tables~\ref{tab:data-full-audit}--\ref{tab:data-scorer-components}).

\subsection{Corpus Characteristics}
\label{sec:data-characteristics}
The \numtotalrecords{} corpus spans broad scenes and themes, uniting urban, architectural, landscape, and indoor Internet clips with interiors, courtyards, mountains, caves, temples, and forests across 14 game/3DGS groups (Appendix~\ref{app:data-coverage}, Table~\ref{tab:data-planning-coverage}).
Trajectories exhibit a global upward trend (median Spearman $\rho=0.916$ game, $0.849$ 3DGS) with permitted local transit dips (${\sim}51\%$ positive adjacent steps; Appendix~\ref{app:data-fixed-time}).

%% file: sec/4_benchmark.tex
\section{Benchmark and Evaluation Infrastructure}
\label{sec:infrastructure}

To evaluate Eq.~\ref{eq:intro-embodied-photo} at minimal cost, we build a 49-scene 3DGS benchmark and a standardized scene conversion interface connecting calibrated assets, closed-loop execution, step budgets, and photographic metrics (Figure~\ref{fig:infrastructure}).
We denote our standard closed-loop evaluation protocol as \textbf{PF41}: from each prescribed starting pose, an agent executes $T=40$ preview-free closed-loop action requests, accounting for \textbf{41 charged photographic observations} in total together with the initial view $I_0$ (Appendix~\ref{app:benchmark}).

\subsection{49-Scene 3DGS Benchmark and Scene Interface}
\label{sec:data-infra}
We adopt lightweight 3DGS for real-time rendering and scalable conversion of real-world scenes into interactive testbeds.
Our 49 benchmark scenes span virtual environments, synthetic objects, and real-world captures (12 game, 15 interior, eight object-centric, four campus/aerial, and ten additional scenes).
Real-world scenes combine dense oblique images and viewpoint photographs for OpenFlyScan reconstruction~\citep{you2026openflyscanqualityguidedaerialreconstruction} with fixed-pose triangulation.
Each standardized scene bundle encapsulates the 3DGS asset, camera calibration, coordinate transform, and collision-free volume (Figure~\ref{fig:data-infra}a), converting new 3DGS reconstructions into plug-and-play testbeds (27.48\,dB mean PSNR on 128 held-out KCD2 views; Appendix~\ref{app:data-infra}).

\subsection{Closed-Loop Interaction and Safe Execution}
\label{sec:infra-observation}\label{sec:infra-execution}\label{sec:infra-records}\label{sec:infra-verification}
Within each standardized scene bundle $\mathcal{W}_e$ (Figure~\ref{fig:infrastructure}b),
world coordinates are right-handed ($+Z$ up) with native metric scale, and the EWA renderer
produces $640\times544$, $90^\circ$ horizontal-FOV, 8-bit RGB observations $I_t$ without
exposing underlying scene geometry or ground-truth poses to the policy.
At step $t$ from pose $T_t=(\mathbf{p}_t, R_t)\in\mathrm{SE}(3)$, action $u_t=(\delta\mathbf{p}_t, \delta R_t)$ updates the pose:
\begin{equation}
 \mathbf{p}_{t+1}=\mathbf{p}_t+R_t\delta\mathbf{p}_t,\qquad R_{t+1}=R_t\delta R_t.
 \label{eq:infra-action}
\end{equation}
To guarantee physical feasibility, the geometric executor verifies that the entire continuous translation segment $[0,1]$ between $\mathbf{p}_t$ and $\mathbf{p}_{t+1}$ remains inside the collision-free volume, preserving $(\mathbf{p}_t, R_t)$ upon rejection.
Under PF41, every action request, including stationary or rejected moves, consumes one step of $T=40$, with poses and RGB hashes logged for deterministic verification across all 49 assets (Appendices~\ref{app:infra-geometry}--\ref{app:infra-cost-views}).

%(evaluated under fast $s_1$ or held-out $s_{2.5}$, $s_{\mathrm{AM}}$, $s_{\mathrm{indep}}$, and $s_{\mathrm{ens}}$)
\subsection{Evaluation Metrics}
\label{sec:infra-eval}
Given the aesthetic score $s(I_t)$ of each observed frame over horizon $T=40$, where $s \in \lbrace s_1,\allowbreak s_{2.5},\allowbreak s_{\mathrm{AM}},\allowbreak s_{\mathrm{indep}},\allowbreak s_{\mathrm{ens}}\rbrace$, we measure closed-loop exploration via best-observed gain $G_{\rm best}$, search efficiency $\operatorname{AUC}$, and success rate $\operatorname{SR}_{0.3}$ ($\mathbb{P}[G_{\rm best}\ge 0.3]$):
\begin{equation}
\begin{aligned}
 G_{\rm best} &= \max_{0\le t\le T} s(I_t) - s(I_0), \\[-2pt]
 \operatorname{AUC} &= \frac{1}{T}\!\sum_{t=1}^{T}\!\Big(\max_{0\le j\le t} s(I_j) - s(I_0)\Big).
\end{aligned}
 \label{eq:task-gains}
\end{equation}
To evaluate onboard view retention across scorers, a Top-$K$ buffer $\mathcal{B}_K$ retains up to $K$ distinct views (\texttt{rgb\_sha256}) ranked by onboard $s_1$ and scores $G_{\rm sel}[s_{\mathrm{indep}}] = \max_{t\in\mathcal{B}_K} s_{\mathrm{indep}}(I_t) - s_{\mathrm{indep}}(I_0)$ ($K=1$ gives $t^\star=\arg\max_t s_1(I_t)$).

%% file: arxiv_sec/5_policy.tex
\begin{figure*}[t]
\centering
\includegraphics[width=0.90\textwidth]{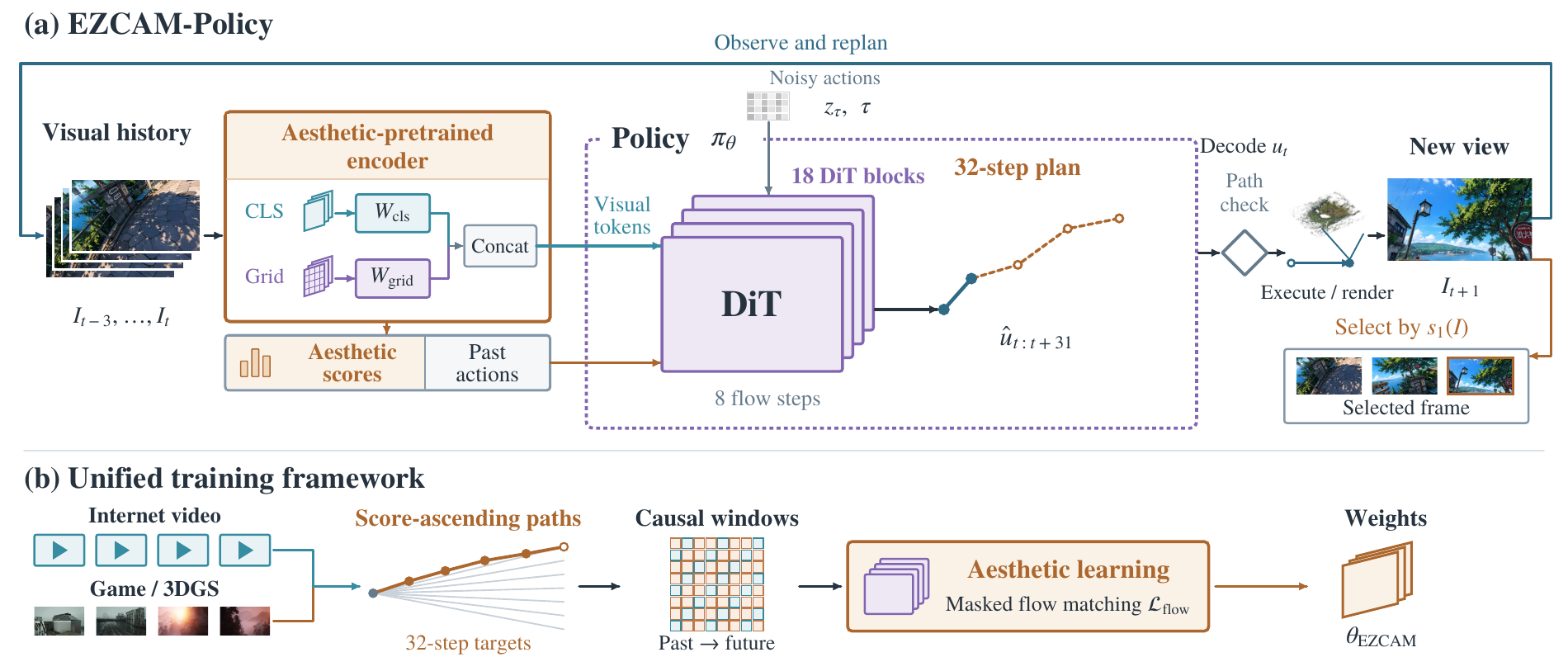}
\caption{\textbf{Architecture and multi-source training of EZCAM-Policy.}
\emph{(a)} The aesthetic-pretrained encoder supplies OpenCLIP tokens and frozen $s_1(I)$ scores to a flow-matching DiT predicting 32-step 6-DoF action chunks.
\emph{(b)} Masked flow matching across Internet, game, and 3DGS data.}
\label{fig:policy-algorithm}
\end{figure*}

\section{Policy and Insights}
\label{sec:method}
\subsection{Insight 1: Why Short-Horizon Behavior Cloning Fails}
As established in Sections~\ref{sec:data-pipelines}--\ref{sec:data-characteristics}, although complete trajectories consistently ascend in aesthetics, intermediate camera motion is locally non-monotonic
(Appendix~\ref{app:data-audit}, Table~\ref{tab:data-full-audit}): only ${\sim}51\%$ of single-step adjacent frames
($t \to t+1$) increase $s(I)$ due to geometric transit and framing adjustments.
In contrast, positive-gain rates climb to \textbf{85.87\%} (game) and \textbf{69.62\%} (3DGS)
over 32-step ($2.13$\,s) action windows, and reach \textbf{96.80\%} and \textbf{92.87\%}
over half-trajectory spans (Appendix~\ref{app:data-fixed-time}, Table~\ref{tab:data-main-windows}).
Consequently, \emph{short-horizon behavior cloning ($k=1$) suffers from local
supervision noise, making long-horizon action prediction ($k=32$) essential to capture
the net trajectory trend of aesthetic improvement} (Table~\ref{tab:policy-horizon}; Appendix~\ref{app:policy-ablations}, Table~\ref{tab:app-b4-stride}).
Beyond mitigating compounding error as in visuomotor action chunking~\citep{zhao2023act,chi2023diffusionpolicy}, $k=32$ chunk prediction here specifically bridges local score non-monotonicity across intermediate framing transit.

\subsection{Insight 2 and EZCAM-Policy Architecture}
Second, because general visual-navigation backbones lack compositional and aesthetic priors,
incorporating \emph{aesthetic pretraining and explicit aesthetic score features} is necessary
for active aesthetic execution.
Guided by both insights, \textbf{EZCAM-Policy} instantiates the policy $\pi_\theta$ in
Eq.~\ref{eq:intro-embodied-photo} by conditioning on $h=4$ RGB observations $I_{t-3:t}$ and
four executed camera actions $u_{t-4:t-1}$.
The \emph{Aesthetic-pretrained encoder} comprises an OpenCLIP ViT-L/14 backbone and frozen LAION V1 CLS heads ($s_1$).
It projects per-frame CLS and $4\times4$ grid tokens ($W_{\rm cls}, W_{\rm grid}$) and supplies them with current-image $s_1$ scores
to an 18-layer, width-1,024 DiT with 16 attention heads (Insight 2),
without using $s_{2.5}$ or $s_{\mathrm{AM}}$.
Rather than relying on noisy single-step prediction ($k=1$), EZCAM-Policy predicts a
long-horizon chunk of $k=32$ camera-local increments $\hat u_{t:t+31}$ (Insight 1), executes
the first decoded increment $u_t$, and replans from $I_{t+1}$, selecting views by $s_1(I)$ (Figure~\ref{fig:policy-algorithm}a).
The deployed policy comprises $927\text{M}$ parameters ($1.78\,\text{GiB}$ FP16 VRAM; full profile in Appendix~\ref{app:real-deployment}).

Each increment $\hat u_{t+i}=(\delta\mathbf p_{t+i},\delta R_{t+i})$ is parameterized by three translation coordinates and the first two
columns of a relative rotation (with a masked 10th padding channel; Appendix~\ref{app:data-training}).
Stored per-channel quantiles $q_{.01,j}$ and $q_{.99,j}$
define the inverse normalization for each action channel $\hat u_{t+i,j}$:
\begin{equation}
 \hat u_{t+i,j}=\frac{\hat z_{t+i,j}+1}{2}(q_{.99,j}-q_{.01,j})+q_{.01,j}.
 \label{eq:policy-decode}
\end{equation}
Decoded actions pass through the same geometric executor as all baselines.

\subsection{Training with Internet, Game, and 3DGS Data}
Internet videos with estimated motion and game/3DGS demonstrations with
recorded poses supply score-ascending windows $(I_{t-h+1:t}, u_{t-h:t-1}, u_{t:t+k-1})$
in the same framework. Inputs contain
past observations $I_{t-h+1:t}$ and executed actions $u_{t-h:t-1}$;
future increments $u_{t:t+k-1}$ are normalized into target chunks $z_0$, with
missing steps masked. Training optimizes the DiT, visual projections, and
final 16 ViT-L blocks while keeping the LAION V1 $s_1$ heads frozen.

For a normalized action-chunk target $z_0$ and conditioning context
$c_t=(I_{t-h+1:t}, u_{t-h:t-1})$, training samples
$z_\tau=(1-\tau)z_0+\tau\epsilon$, with flow time $\tau\sim U(0,1)$ and Gaussian noise $\epsilon$.
Velocity $v_\theta(z_\tau,\tau\mid c_t)$ is trained via masked flow matching (Figure~\ref{fig:policy-algorithm}b):
\begin{equation}
 \mathcal L_{\rm flow}=\mathbb E\!\left[
 \frac{\|M\odot(v_\theta(z_\tau,\tau\mid c_t)- (\epsilon-z_0))\|_2^2}{\|M\|_1}
 \right],
 \label{eq:policy-flow}
\end{equation}
where $M$ selects valid future translation and rotation channels across the $k=32$ horizon.
Eight decoding steps produce the predicted action chunk $\hat u_{t:t+k-1}$. Executed motion, including
rejections, updates the action history $u_{t-h+1:t}$; every request consumes interaction budget.

\paragraph{Branch-selected continuation targets.}
On game and 3DGS data, EZCAM-Policy does not merely imitate a single recorded window.
When multiple score-ascending graph paths branch from a shared 3D region, branch-selected continuation supervises $\hat u_{t:t+31}$ with the highest-return outgoing continuation while excluding future branch images from the encoder, teaching $\pi_\theta$ to resolve multi-modal junctions toward peak compositions.
The frozen LAION V1 $s_1$ head supplies fast online score conditioning without accessing $s_{2.5}$ or $s_{\mathrm{AM}}$; both components are ablated in Section~\ref{sec:ablations} (Appendix~\ref{app:policy-ablations}).

%% file: arxiv_sec/6_experiments.tex
\begin{table*}[t]
\centering\footnotesize
\setlength{\tabcolsep}{2.9pt}
\caption{\textbf{Main closed-loop comparison across all 49 worlds ($n=294$ episodes per model, $T=40$; $2\text{k}$ checkpoint).}
Six shared starts per scene under PF41, reporting primary $s_1(I)$ metrics and held-out LAION-free $G_{\rm best}[s_{\mathrm{indep}}]$ on Unseen 36 ($\operatorname{SR}_{0.3}$ in $\%$). \textbf{EZCAM-Policy} leads across every split, scorer, and domain.}
\label{tab:policy-main}\label{tab:policy-scene-splits}\label{tab:policy-domains}
\resizebox{\textwidth}{!}{%
\begin{tabular}{@{}lrrrrrrrrrrrrrrrr@{}}\toprule
& \multicolumn{3}{c}{All Worlds (49)}
& \multicolumn{5}{c}{Exposure Splits (Seen 10 / Dev.\ 3 / Unseen 36)}
& \multicolumn{2}{c}{Game Env.\ (12)}
& \multicolumn{2}{c}{Real Indoor (15)}
& \multicolumn{2}{c}{Campus/Aerial (4)}
& \multicolumn{2}{c}{Landmarks/Obj.\ (18)}\\
\cmidrule(lr){2-4}\cmidrule(lr){5-9}\cmidrule(lr){10-11}\cmidrule(lr){12-13}\cmidrule(lr){14-15}\cmidrule(l){16-17}
Model & $G_{\rm best}\uparrow$ & AUC$\uparrow$ & SR$_{0.3}\uparrow$
& Seen $G_{\rm best}$ & Dev.\ $G_{\rm best}$ & Unseen $G_{\rm best}$ & Unseen SR$_{0.3}$ & Unseen $G_{\rm best}[s_{\mathrm{indep}}]$
& $G_{\rm best}$ & SR$_{0.3}$ & $G_{\rm best}$ & SR$_{0.3}$ & $G_{\rm best}$ & SR$_{0.3}$ & $G_{\rm best}$ & SR$_{0.3}$\\\midrule
\textbf{EZCAM-Policy (ours)} & \textbf{0.5126} & \textbf{0.3857} & \textbf{57.8} & \textbf{0.8480} & \textbf{0.7638} & \textbf{0.3985} & \textbf{47.2} & \textbf{0.3758} & \textbf{0.8396} & \textbf{88.9} & \textbf{0.3986} & \textbf{50.0} & \textbf{0.6306} & \textbf{75.0} & \textbf{0.3633} & \textbf{39.8} \\
DualVLN~\citep{wei2025dualvln} & 0.2817 & 0.2438 & 30.6 & 0.7919 & 0.6483 & 0.1094 & 13.9 & 0.0927 & 0.7882 & 77.8 & 0.0791 & 8.9 & 0.3013 & 45.8 & 0.1085 & 13.9 \\
StreamVLN~\citep{wei2025streamvln} & 0.2637 & 0.2159 & 32.0 & 0.4167 & 0.4538 & 0.2054 & 23.6 & 0.2193 & 0.4165 & 54.2 & 0.1564 & 17.8 & 0.4285 & 54.2 & 0.2147 & 24.1 \\
Uni-NaVid~\citep{zhang2025uninavid} & 0.2302 & 0.1861 & 31.6 & 0.4087 & 0.4755 & 0.1602 & 21.3 & 0.1614 & 0.4267 & 61.1 & 0.1077 & 13.3 & 0.3237 & 54.2 & 0.1806 & 22.2 \\
NavGPT-2~\citep{zhou2024navgpt2} & 0.2171 & 0.1841 & 27.6 & 0.5416 & 0.5425 & 0.0999 & 12.0 & 0.1256 & 0.5612 & 72.2 & 0.0589 & 6.7 & 0.3388 & 50.0 & 0.0926 & 10.2 \\
NaviLLM~\citep{zheng2024navillm} & 0.2077 & 0.1755 & 27.9 & 0.3703 & 0.5440 & 0.1345 & 15.7 & 0.1537 & 0.3870 & 61.1 & 0.1047 & 10.0 & 0.3821 & 50.0 & 0.1352 & 15.7 \\
NaVid~\citep{zhang2024navid} & 0.2006 & 0.1679 & 23.5 & 0.3373 & 0.4895 & 0.1386 & 13.9 & 0.1482 & 0.3644 & 48.6 & 0.0861 & 6.7 & 0.3668 & 41.7 & 0.1500 & 16.7 \\\midrule
Yaw scan & 0.2365 & 0.2189 & 31.0 & 0.3227 & 0.3061 & 0.2068 & 26.4 & 0.2245 & 0.3200 & 43.1 & 0.2224 & 28.9 & 0.2649 & 33.3 & 0.1864 & 24.1 \\
Local random walk & 0.1199 & 0.0735 & 13.6 & 0.2426 & 0.0725 & 0.0898 & 11.1 & 0.1430 & 0.2202 & 22.2 & 0.0599 & 4.4 & 0.0917 & 8.3 & 0.1094 & 16.7 \\
\bottomrule\end{tabular}%
}
\vspace{-5pt}
\end{table*}

\begin{figure*}[t]
\centering
\includegraphics[width=0.96\textwidth]{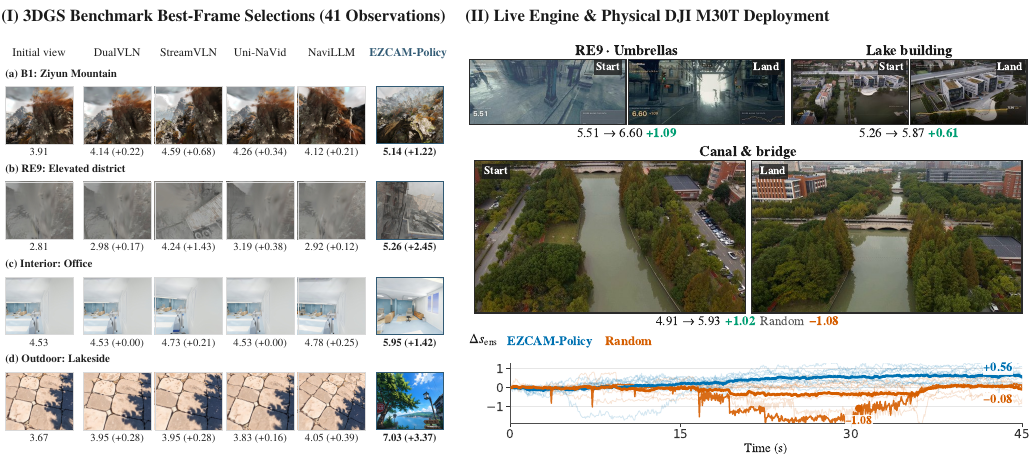}
\caption{\textbf{Qualitative closed-loop framing and zero-shot physical deployment.}
\emph{(I)} Best-observed views within $T=40$ steps from identical starts ($s_{\mathrm{ens}}(I)$ and $\Delta s_{\mathrm{ens}}$ below each frame).
\emph{(II)} Zero-shot closed-loop execution in live game engines and onboard a DJI Matrice 30T drone with continuous flight traces (Appendix~\ref{app:real-deployment}, Figure~\ref{fig:real-deployment}).}
\label{fig:policy-best-frames}
\end{figure*}

\section{Experiments}
\label{sec:experiments}
\subsection{Experimental Setup and Main Closed-Loop Comparison}
\label{sec:policy-eval}
\begin{figure}[!htb]
\centering
\begin{minipage}[b]{0.49\textwidth}
  \centering
  \includegraphics[width=\linewidth]{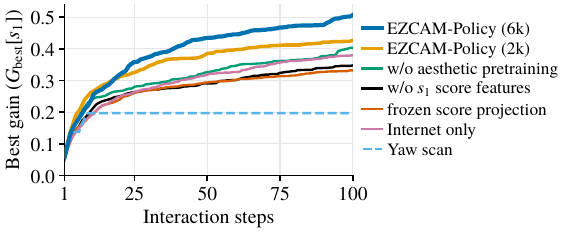}\\[-2pt]
  {\footnotesize\textbf{(a)} Policy ablations and $2\text{k}$ vs.\ $6\text{k}$ ($T=1\dots 100$)}
\end{minipage}\hfill
\begin{minipage}[b]{0.49\textwidth}
  \centering
  \includegraphics[width=\linewidth]{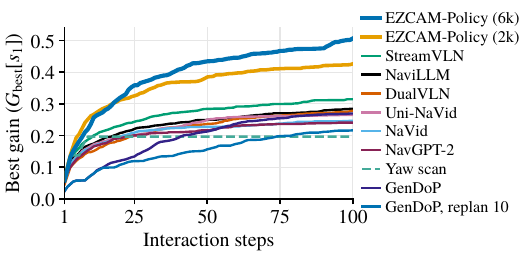}\\[-2pt]
  {\footnotesize\textbf{(b)} External baselines and $2\text{k}$ vs.\ $6\text{k}$ ($T=1\dots 100$)}
\end{minipage}
\caption{\textbf{Long-horizon budget scaling on 30 unseen scenes ($n=180$, $T=1\dots 100$).}
\emph{(a)} Matched $2\text{k}$ ablations paired with full-epoch $6\text{k}$ training.
\emph{(b)} Both \textbf{EZCAM-Policy ($2\text{k}$)} and \textbf{($6\text{k}$)} outperform all baselines across budgets, with full-epoch $6\text{k}$ lifting long-horizon scaling.}
\label{fig:policy-unseen30-external}\label{fig:policy-unseen30-ablations}
\end{figure}

We evaluate all methods under PF41 across six shared, aesthetically uncurated starts per scene ($n=294$, $T=40$; Appendix~\ref{app:infra-verification}, Table~\ref{tab:infra-verification-summary}).
Here, $6\text{k}$ ($6{,}000$ updates) corresponds to one full training epoch, whereas $2\text{k}$ ($2{,}000$ updates) already converges sufficiently to demonstrate our core methodological gains; to balance compute across extensive ablations and evaluations, we adopt the $2\text{k}$ checkpoint throughout main comparisons, ablations, and deployments, and pair $6\text{k}$ alongside $2\text{k}$ in long-horizon scaling (Section~\ref{sec:multi-budget}, Figure~\ref{fig:policy-unseen30-external}).
Table~\ref{tab:policy-main} compares \textbf{EZCAM-Policy} against six VLN backbones adapted with 10-D action heads on EZCAM data (DualVLN~\citep{wei2025dualvln}, StreamVLN~\citep{wei2025streamvln}, Uni-NaVid~\citep{zhang2025uninavid}, NavGPT-2~\citep{zhou2024navgpt2}, NaviLLM~\citep{zheng2024navillm}, NaVid~\citep{zhang2024navid}; Appendix~\ref{app:adapter-results}) and two zero-inference heuristics (Yaw scan, Local random walk; Appendix~\ref{app:motion-baselines}), with qualitative views in Figure~\ref{fig:policy-best-frames}\,I and per-scene margins plus capture-pose independence checks in Appendix~\ref{app:policy-comparison} (Figure~\ref{fig:policy-full49}).
Lacking compositional priors, VLN adapters suffer deterministic collision stalls on cluttered unseen indoor scenes and fall below in-place Yaw scan, while Yaw scan cannot change 3D parallax and Local random walk drifts aimlessly; \textbf{EZCAM-Policy} actively maneuvers around obstacles toward stronger compositions, leading every baseline across all splits, held-out unseen evaluation, and domains (Table~\ref{tab:policy-main}).

\begin{table}[!htb]
\centering\small
\setlength{\tabcolsep}{7.5pt}
\renewcommand{\arraystretch}{1.10}
\caption{\textbf{Dynamic Top-$K$ shutter selection under held-out $s_{\mathrm{indep}}$ ($T=40$; $2\text{k}$).} $G_{\rm sel}[s_{\mathrm{indep}}]$ for top-$K$ distinct views ranked by $s_1$ vs.\ 41-frame peak across same-protocol PF41 splits (Appendix~\ref{app:heldout-shutter}).}
\label{tab:policy-heldout-shutter}
\resizebox{\linewidth}{!}{%
\begin{tabular}{@{}llrrrrr@{}}\toprule
Split & Method & $K{=}1$ & $K{=}3$ & $K{=}5$ & $K{=}10$ & 41-fr.\ Peak \\\midrule
\multirow{4}{*}{\shortstack[l]{All Worlds 49\\($n=294$)}}
& \textbf{EZCAM-Policy} & \textbf{0.1039} & \textbf{0.2716} & \textbf{0.3533} & \textbf{0.4360} & \textbf{0.5213} \\
& StreamVLN             & 0.0531          & 0.1842          & 0.2240          & 0.2760          & 0.3550          \\
& DualVLN               & 0.0258          & 0.1521          & 0.1740          & 0.2141          & 0.2599          \\
& Yaw scan              & 0.0263          & 0.1965          & 0.2401          & 0.2658          & 0.2658          \\\midrule
\multirow{4}{*}{\shortstack[l]{Unseen 36\\($n=216$)}}
& \textbf{EZCAM-Policy} & \textbf{0.0533} & \textbf{0.2025} & \textbf{0.2735} & \textbf{0.3281} & \textbf{0.3758} \\
& StreamVLN             & 0.0112          & 0.1173          & 0.1447          & 0.1681          & 0.2193          \\
& DualVLN               & $-0.0330$       & 0.0637          & 0.0716          & 0.0867          & 0.0927          \\
& Yaw scan              & 0.0233          & 0.1618          & 0.2062          & 0.2245          & 0.2245          \\\midrule
\multirow{4}{*}{\shortstack[l]{Unseen 30\\($n=180$)}}
& \textbf{EZCAM-Policy} & \textbf{0.0677} & \textbf{0.2258} & \textbf{0.3046} & \textbf{0.3677} & \textbf{0.4241} \\
& StreamVLN             & 0.0216          & 0.1346          & 0.1641          & 0.1908          & 0.2516          \\
& DualVLN               & $-0.0297$       & 0.0689          & 0.0781          & 0.0962          & 0.1034          \\
& Yaw scan              & 0.0348          & 0.1768          & 0.2273          & 0.2452          & 0.2452          \\\bottomrule
\end{tabular}%
}
\end{table}

\paragraph{Held-out scorers and Top-$K$ onboard shutter.}
Since data filtering and internal policy conditioning use only $s_1$ (Sections~\ref{sec:data-scorers} and~\ref{sec:method}), $s_{2.5}$ and $s_{\mathrm{AM}}$ remain strictly unseen during training.
Deterministic replay confirms that \textbf{EZCAM-Policy} leads across all held-out evaluators ($s_{2.5}$, $s_{\mathrm{AM}}$, and $s_{\mathrm{indep}}$; Table~\ref{tab:policy-main} and Appendix~\ref{app:heldout-shutter}, Tables~\ref{tab:app-a1-all49}--\ref{tab:app-a1-unseen30-strict}) and in mean gain from $I_0$ to the best-observed frame under third-party IQA models Q-Align~\citep{wu2024qalign}, MUSIQ~\citep{ke2021musiq}, and NIMA~\citep{talebi2018nima} (Appendix~\ref{app:external-iqa}, Table~\ref{tab:app-b3-iqa}).
Moreover, via its built-in onboard aesthetic module, \textbf{EZCAM-Policy} dynamically selects and retains the top-$K$ highest-scoring distinct views ($K\in\{1,3,5,10\}$) as shutter outputs during exploration, an active shutter capability absent in pure navigation baselines that execute path movement without aesthetic perception.
Even when baselines are retroactively granted the identical $s_1$-ranked distinct-view buffer, \textbf{EZCAM-Policy} achieves strong independent gains even at tight quotas (top 3 or 5 views), leading every baseline across all $K$ and rapidly approaching the full-trajectory peak (Table~\ref{tab:policy-heldout-shutter}).

\paragraph{Blinded human evaluation on policy outputs.}
In a double-blind 2AFC study ($1{,}590$ judgments across 15 scene pairs), \textbf{106 participants} compared \textbf{EZCAM-Policy} against the oracle \textbf{per-scene best baseline} (\emph{Best-of-Baselines}) from shared starts (Appendix~\ref{app:policy-human}, Figure~\ref{fig:app-policy-human-106} and Table~\ref{tab:policy-human-106}), overwhelmingly preferring \textbf{EZCAM-Policy} ($p<10^{-10}$) across every scene pair and virtually all participants without positional bias.

\subsection{Multi-Budget Unseen Generalization}
\label{sec:multi-budget}
Scaling up to $T=100$ steps on 30 unseen scenes (30-scene diagnostic, $n=180$), we pair our default $2\text{k}$ checkpoint alongside the full-epoch $6\text{k}$ checkpoint against GenDoP~\citep{zhang2025gendop}, the six VLN adapters, and motion heuristics (Figure~\ref{fig:policy-unseen30-external}a,b; Appendix~\ref{app:policy-ablations}, Tables~\ref{tab:policy-unseen30-k100}--\ref{tab:policy-ablations}).
In-place Yaw scan saturates once its local angular sweep completes, non-interactive trajectory generation (with or without periodic open-loop replanning) cannot adaptively correct framing against unseen 3D scene geometry, and VLN adapters plateau early once stalled near local obstacles.
By contrast, closed-loop receding-horizon control enables both \textbf{EZCAM-Policy ($2\text{k}$)} and full-epoch \textbf{($6\text{k}$)} to continuously discover stronger viewpoints as interaction budgets grow, with full-epoch $6\text{k}$ training further raising the long-horizon ceiling (Figure~\ref{fig:policy-unseen30-external}b).

\subsection{Ablation on Multi-Source Training Data}
\label{sec:ablations-data}
Table~\ref{tab:policy-data} ablates $2\text{k}$ training sources across all 49 worlds ($T=40$).
Metrically calibrated Game and 3DGS trajectories provide exact 6-DoF action-to-viewpoint supervision that drives rapid early search efficiency ($\operatorname{AUC}$), whereas Internet clips contribute broad scene semantics and real-world framing diversity that elevate overall completion ($\operatorname{SR}_{0.3}$).
Combining both unites metric 3D controllability with open-world visual generalization, achieving the strongest performance across all metrics on both the 49-scene benchmark (Table~\ref{tab:policy-data}) and unseen scaling (Figure~\ref{fig:policy-unseen30-ablations}a).

\begin{table}[!htb]
\centering\small
\setlength{\tabcolsep}{10pt}
\renewcommand{\arraystretch}{1.12}
\caption{\textbf{Multi-source data ablation ($n=294$, $T=40$; $2\text{k}$).}}
\label{tab:policy-data}
\begin{tabular}{@{}ccrrr@{}}\toprule
Game/3DGS & Internet & $G_{\rm best}\uparrow$ & $\operatorname{AUC}\uparrow$ & $\operatorname{SR}_{0.3}$(\%)$\uparrow$\\\midrule
$\checkmark$ & --- & 0.4453 & 0.3670 & 46.3\\
--- & $\checkmark$ & 0.4169 & 0.2926 & \textbf{57.8}\\
$\checkmark$ & $\checkmark$ & \textbf{0.5126} & \textbf{0.3857} & \textbf{57.8}\\\bottomrule
\end{tabular}
\end{table}

\subsection{Ablation on Policy Insights and Architecture}
\label{sec:ablations}

\begin{table}[!htb]
\centering\small
\setlength{\tabcolsep}{8pt}
\renewcommand{\arraystretch}{1.12}
\caption{\textbf{Prediction horizon sweep ($n=294$, $T=40$; single-stage $2\text{k}$).}}
\label{tab:policy-horizon}
\resizebox{\linewidth}{!}{%
\begin{tabular}{@{}lrrrr@{}}\toprule
Horizon $k$ & $G_{\rm best}\uparrow$ & $\operatorname{AUC}\uparrow$ & $\operatorname{SR}_{0.3}$(\%)$\uparrow$ & $\Delta G_{\rm best}$ vs.\ $k=32$ [95\% CI]\\\midrule
\textbf{$k=32$ (ours)} & \textbf{0.5711} & \textbf{0.4116} & \textbf{65.0} & Reference\\
$k=16$ & 0.4636 & 0.3493 & 50.0 & $-0.1074\;[-0.1779, -0.0402]$\\
$k=8$ & 0.3142 & 0.2328 & 37.4 & $-0.2569\;[-0.3204, -0.1938]$\\
$k=4$ & 0.4340 & 0.3261 & 52.4 & $-0.1371\;[-0.2179, -0.0640]$\\
$k=2$ & 0.3855 & 0.2769 & 44.9 & $-0.1856\;[-0.2605, -0.1183]$\\
$k=1$ (short BC) & 0.4549 & 0.3179 & 50.7 & $-0.1162\;[-0.1775, -0.0579]$\\\bottomrule
\end{tabular}%
}
\end{table}

Unlike the default two-stage $2\text{k}$ policy evaluated in Tables~\ref{tab:policy-main}--\ref{tab:policy-data}, Table~\ref{tab:policy-horizon} trains all horizons $k\in\{1,2,4,8,16,32\}$ in a single $2\text{k}$ stage on a unified 32-step window cache to isolate \textbf{Insight 1}.
While intermediate horizons exhibit single-seed optimization variance because truncated windows can terminate mid-transit inside local aesthetic valleys, long-horizon $k=32$ chunk prediction significantly beats single-step behavior cloning ($k=1$) and every shorter horizon across all metrics (Table~\ref{tab:policy-horizon}), confirming that multi-step targets bridge local score non-monotonicity.
Sweeping test-time execution stride $m\in\{1,2,4,8,16,32\}$ on the frozen default $2\text{k}$ checkpoint further decouples chunk supervision from reactive replanning (Appendix~\ref{app:policy-ablations}, Figure~\ref{fig:app-b4-stride} and Table~\ref{tab:app-b4-stride}): executing the full 32-step chunk ($m=32$) matches reactive $m=1$ across all worlds and surpasses it on unseen scenes, while moderate strides ($m=4$ or $m=8$) lift peak gain and cut inference calls by $4\text{--}8\times$ by suppressing step-to-step resampling jitter.
Matched $2\text{k}$ architectural ablations (Figure~\ref{fig:policy-unseen30-ablations}a; Appendix~\ref{app:policy-ablations}, Tables~\ref{tab:policy-ablation-49} and~\ref{tab:policy-ablations}) validate \textbf{Insight 2} and branch-selected continuation: removing aesthetic pretraining, explicit $s_1$ conditioning, learnable score projection, or highest-return branch continuation consistently degrades both search efficiency and peak composition quality.

\subsection{Live Engine and Drone Deployment}
\label{sec:real-deployment}
We deploy \textbf{EZCAM-Policy} ($2\text{k}$) zero-shot in live game engines (\emph{KCD2}, \emph{Black Myth: Wukong}, and \emph{RE9}; 16 episodes) and onboard a physical \textbf{DJI Matrice 30T} quadrotor (\textbf{NVIDIA Orin NX}) across five multi-scale aerial paths and 35 controlled outdoor flights ($36.3$\,min, $32{,}680$ frames) against an onboard random policy using only ego-centric RGB history.
As shown in Figure~\ref{fig:policy-best-frames}\,II and Table~\ref{tab:real-world-deployment} (full per-episode logs and hardware telemetry in Appendix~\ref{app:real-deployment}, Figure~\ref{fig:real-deployment} and Tables~\ref{tab:suppl-live-games}--\ref{tab:suppl-real-system}), \textbf{EZCAM-Policy} runs with sub-second onboard latency and accurate 6-DoF gimbal--airframe tracking, improves framing across every live-engine run, and significantly outperforms random exploration in outdoor flights ($p=6.4\times 10^{-5}$; $p<10^{-4}$ after ANCOVA adjustment for initial score $s_{\mathrm{ens}}(I_0)$), confirming map-free sim-to-real transfer under natural lighting and wind.

\begin{table}[!htb]
\centering\small
\setlength{\tabcolsep}{7pt}
\renewcommand{\arraystretch}{1.12}
\caption{\textbf{Live-engine and DJI M30T deployment ($2\text{k}$).} Rows 1--2: episode peaks; rows 3--6: controlled outdoor flights ($G_{\rm best}@60$\,s peaks and $\Delta s_{\mathrm{ens}}@30$\,s fixed-horizon endpoints).}
\label{tab:real-world-deployment}
\resizebox{\linewidth}{!}{%
\begin{tabular}{@{}llrrrr@{}}\toprule
Deployment Domain & Policy & Runs & Start $s_{\mathrm{ens}}(I_0)$ & Peak / End $s_{\mathrm{ens}}(I)$ & Gain $\Delta s_{\mathrm{ens}}\uparrow$\\\midrule
Live Game Engines (KCD2, Wukong, RE9) & \textbf{EZCAM-Policy} & 16 & 4.70 & 6.05 & \textbf{+1.35}\\
DJI Matrice 30T (Multi-Scale Aerial) & \textbf{EZCAM-Policy} & 5 & 5.08 & 5.76 & \textbf{+0.68}\\\midrule
DJI M30T Controlled ($G_{\rm best}@60$\,s) & \textbf{EZCAM-Policy} & 16 & 4.94 & 5.66 & \textbf{+0.73}\\
DJI M30T Controlled ($G_{\rm best}@60$\,s) & Random Policy & 17 & 5.26 & 5.48 & +0.22\\\midrule
DJI M30T Controlled ($\Delta s_{\mathrm{ens}}@30$\,s) & \textbf{EZCAM-Policy} & 13 & 4.86 & 5.32 & \textbf{+0.46}\\
DJI M30T Controlled ($\Delta s_{\mathrm{ens}}@30$\,s) & Random Policy & 10 & 5.33 & 4.96 & $-0.37$\\\bottomrule
\end{tabular}%
}
\end{table}

%% file: sec/7_conclusion.tex
\section{Conclusion and Limitations}
\label{sec:conclusion}\label{sec:limitations}
We presented EZCAM to establish embodied photography as a closed-loop sequential decision-making problem across data, benchmark, and policy learning, uniting \numtotalrecords{} score-ascending records with a 49-scene 3DGS interactive testbed and standardized scene interface.
Grounded in our insights on long-horizon action-chunk prediction across non-monotonic transit and aesthetic pretraining with onboard shutter selection, \textbf{EZCAM-Policy} outperforms navigation, trajectory-generation, and heuristic baselines across held-out evaluators, wins \textbf{93.8\%} of blinded human judgments, and transfers zero-shot to live game engines and physical DJI Matrice 30T drones.
While EZCAM currently assumes static scenes and fixed camera intrinsics, extending closed-loop framing to dynamic subjects and joint focal, aperture, exposure, and 6-DoF control are promising directions.
\label{zz:end-main-body}%

%% file: arxiv_sec/X_suppl.tex
\ifnum\value{page}>1
\clearpage
\setcounter{page}{1}
\ifdefined\hypersetup
\hypersetup{pageanchor=false}
\fi
\fi
\maketitlesupplementary
\appendix
\renewcommand{\thetable}{S\arabic{table}}
\renewcommand{\thefigure}{S\arabic{figure}}
\renewcommand{\theequation}{S\arabic{equation}}
\ifdefined\theHtable
\renewcommand{\theHtable}{S\arabic{table}}
\renewcommand{\theHfigure}{S\arabic{figure}}
\renewcommand{\theHequation}{S\arabic{equation}}
\fi
\setcounter{table}{0}
\setcounter{figure}{0}
\setcounter{equation}{0}

\section{Data construction details}
\label{app:data}

This supplement presents full-scale verification of the EZCAM data engine, benchmark, and policy across 44.3M multi-scorer trajectory frames, 194,070 deterministically replayed benchmark views, 3,810 blinded human preference judgments, and live commercial game-engine and physical DJI Matrice 30T drone deployments.

\subsection{Scoring-model documentation}
\label{app:data-models}

To instantiate the multi-backbone evaluation criterion in Eq.~\ref{eq:intro-embodied-photo}, we combine three complementary aesthetic predictors: the fast two-head LAION V1 predictor~\citep{schuhmann2022laionaesthetics} $s_1(I)$ (averaging OpenCLIP ViT-L/14 and ViT-B/32 heads, native $[0,10]$ scale), the $[0,10]$-aligned SigLIP-SO400M V2.5 predictor~\citep{discus0434aestheticv25} $s_{2.5}(I)=\operatorname{clip}(s_{2.5}^{\mathrm{raw}}(I),0,10)$ (averaging its two heads), and the $[0,10]$-aligned MLLM ArtiMuse predictor~\citep{cao2025artimuse} $s_{\mathrm{AM}}(I)=s_{\mathrm{AM}}^{\mathrm{raw}}(I)/10$.
Their three-model equal-weight ensemble is
\begin{equation}
 s_{\mathrm{ens}}(I)=\frac{1}{3}\Big[
 s_1(I)+s_{2.5}(I)
 +s_{\mathrm{AM}}(I)\Big].
 \label{eq:scorer-normalized-mean}
\end{equation}
Dividing ArtiMuse's $0\text{--}100$ output by ten and clipping V2.5 to $[0,10]$ align all three predictors to common nominal $0\text{--}10$ units, while fixed equal $1/3$ weights reproducibly unite three complementary visual architectures (OpenCLIP regression heads, SigLIP, and expert-annotated MLLM assessment) into a calibration-free multi-model criterion without fitting post-hoc weights on benchmark data.
High-throughput data curation and real-time onboard policy conditioning leverage fast two-head LAION V1 ($s_1$) for scalable scoring across tens of millions of candidate views, while full-corpus trajectory validation, live/real-world deployments, and deterministic replay cross-scorer evaluations (Section~\ref{sec:policy-eval}, Appendix~\ref{app:heldout-shutter}) employ all three models via $s_{\mathrm{ens}}(I)$ in Eq.~\ref{eq:scorer-normalized-mean} and the strictly LAION-free ensemble $s_{\mathrm{indep}}(I)=\frac{1}{2}[s_{2.5}(I)+s_{\mathrm{AM}}(I)]$, keeping $s_{2.5}$ and $s_{\mathrm{AM}}$ completely unseen during data collection and policy training.
In our blinded human study ($N=37$ participants over 60 pairs: 21 game, 39 3DGS; 2,220 responses comparing initial views against top views within 10\,s), higher-$s_{\mathrm{ens}}(I)$ images won 1,417 of 1,609 directional preferences (\textbf{88.07\%}; two-way 95\% bootstrap CI: 83.2--92.2\%; $s_1$ alone: \textbf{87.88\%}), as detailed in Appendices~\ref{app:data-human}--\ref{app:human-components}.

\subsection{Corpus inventory and record units}
\label{app:data-records}

\begin{table}[!htb]
 \centering\renewcommand{\arraystretch}{1.08}
 \small
 \caption{\textbf{Corpus inventory and planning-pose coverage.} Pose and endpoint counts are distinct within each group before summing (per-group breakdown in Table~\ref{tab:data-planning-coverage}).}
 \label{tab:data-main-sources}\label{tab:data-sources}
 \resizebox{\linewidth}{!}{%
 \begin{tabular}{@{}lrrrr@{}}
 \toprule
 Source & Records & Groups & Key poses & Endpoints \\
 \midrule
 Game engine & 61,000 & 7 & 17,172 & 10 \\
 Game-derived 3DGS & 35,700 & 7 & 24,072 & 9 \\
 Internet (clips) & 347,111 & --- & --- & --- \\
 \midrule
 Total & 443,811 & 14 & 41,244 & 19 \\
 \bottomrule
 \end{tabular}%
 }
\end{table}

\paragraph{Record units and provenance.}
Collection groups capture multi-trajectory coverage across distinct regions or temporal configurations within each source environment.
Scene-level train/test splitting strictly follows original environment identities: engine scenes and their reconstructions share that identity.
Game poses come from engine logs, 3DGS poses from rendering cameras, and
Internet poses are estimated up to relative scale.
Each trajectory retains video $V_\tau$, camera poses $P_\tau$, and scored
planning views:
\begin{equation}
 \begin{aligned}
 \mathcal R_\tau &= (V_\tau,P_\tau,\mathcal K_\tau),\\
 \mathcal K_\tau &= ((q_{\tau k},I_{\tau k},c_{\tau k}))_{k=1}^{K_\tau}.
 \end{aligned}
 \label{eq:data-record}
\end{equation}
Here $c_{\tau k}=s_1(I_{\tau k})$ is the lightweight LAION V1 collection score of image $I_{\tau k}$ at pose $q_{\tau k}$.
Decoded frames, recorded poses, planning-view occurrences, and training
windows are separate units; views can recur across trajectories.

\subsection{View acquisition, path construction, and replay}
\label{app:data-acquisition}
\label{sec:data-construction}

\paragraph{Acquisition.}
Our self-developed general-purpose UE5 capture platform, compatible with virtually all Unreal Engine titles alongside game-specific adapters, provides
camera control, pose readout, scanning, and recording
(Figure~\ref{fig:data-keyview-examples}a). The viewpoint map
$\mathcal M_e=\{(q_j,I_j,c_j)\}_{j=1}^{N_e}$ associates poses, images, and
scores. Reconstructed 3DGS environments~\citep{kerbl20233dgs} are rendered
and scored separately before path search.

The base scan samples 22 directions per safe-grid position: eight yaw
angles at $45^\circ$ spacing and zero pitch, six at $60^\circ$ spacing
for each of pitch $\pm45^\circ$, and two vertical views. Supplementary
captures add candidate poses.

\paragraph{Directed path search.}
The viewpoint map supplies nodes for a directed graph $G_e=(V_e,E_e)$.
Let $\mathbf{p}_i$ be camera position, $c_i$ its recorded aesthetic
score, $\delta_\psi(i,j)$ the absolute shortest yaw difference, and
$\delta_\theta(i,j)$ the absolute pitch difference. Edges satisfy
\begin{equation}
 (i,j)\in E_e\;\Longleftrightarrow\;
 \begin{cases}
 d_{\min}\leq\|\mathbf{p}_j-\mathbf{p}_i\|_2\leq d_{\max},\\
 \delta_\psi(i,j)\leq\alpha_{\max},\;
 \delta_\theta(i,j)\leq\beta_{\max},\\
 0<c_j-c_i\leq\Delta c_{\max}.
 \end{cases}
 \label{eq:data-edges}
\end{equation}
Retained sequences have positive minimum key-view increment:
\begin{equation}
 m_\tau=\min_{1\leq k<K_\tau}
       (c_{\tau,k+1}-c_{\tau k})>0.
 \label{eq:data-anchor-order}
\end{equation}

\paragraph{Geometric path cost.}
For $K_\tau$ key poses, the path cost combines translation, rotation, and
changes in travel direction:
\begin{equation}
 \begin{split}
 J(\tau)={}&\sum_{k=1}^{K_\tau-1}\big[
 \|\mathbf{p}_{k+1}-\mathbf{p}_k\|_2
 +\lambda_\psi\,\delta_\psi(k,k+1) \\
 &\qquad\quad +\lambda_\theta\,\delta_\theta(k,k+1)\big]
 +\lambda_{\mathrm{turn}}\sum_{k=2}^{K_\tau-1}\varphi_k,
 \end{split}
 \label{eq:data-path-cost}
\end{equation}
where $\varphi_k$ is the angle between consecutive displacement vectors and
the $\lambda$ coefficients weight the angular costs. Search also constrains
geometric revisits and path turns; score ordering follows
Eq.~\ref{eq:data-edges}.

\paragraph{Time parameterization.}
Let $t_k$ be the time assigned to the $k$th key pose. The replay export sets
$t_1=0$ and $t_{k+1}=t_k+\Delta t_k$, with
\begin{equation}
 \begin{aligned}
 \Delta t_k=\max\Bigl\{
 &t_{\min},\,
 \frac{\|\mathbf{p}_{k+1}-\mathbf{p}_k\|_2}{v_p},\\
 &\frac{\delta_\psi(k,k+1)}{v_\psi},\,
 \frac{\delta_\theta(k,k+1)}{v_\theta}
 \Bigr\}.
 \end{aligned}
 \label{eq:data-timing}
\end{equation}
Here $t_{\min}$ is the minimum duration; $v_p$, $v_\psi$, and
$v_\theta$ are translation/yaw/pitch rates in matching native units.

\paragraph{Pose interpolation and screening.}
Position and viewing angles use cubic Hermite interpolation:
\begin{equation}
 \begin{aligned}
 x(w)={}&h_{00}(w)x_k+h_{10}(w)\Delta t_k d_k\\
        &+h_{01}(w)x_{k+1}+h_{11}(w)\Delta t_k d_{k+1},
 \end{aligned}
 \label{eq:data-hermite}
\end{equation}
where $h_{ij}$ are the standard Hermite basis functions, $d_k$ are endpoint slopes, and
$w=(t-t_k)/\Delta t_k$. Positional derivatives use damped finite differences
with zero endpoints; unwrapped yaw and pitch use shape-preserving slopes.
Sampling includes the exact key poses and their times.
Game and 3DGS replay enforce safe-volume and depth screening along the executed path: game replay uses safe regions and depth tests, and 3DGS replay uses safe positions and reconstructed geometry.
Continuous Hermite interpolation naturally traverses the intermediate geometric transition poses between high-scoring key views.
That transition structure motivates the multi-horizon temporal analysis in Appendix~\ref{app:data-fixed-time} and long-horizon chunk prediction.

\paragraph{Recording configurations and coverage.}

Game recordings comprise 52,259 videos at 30\,FPS, 5,741 at 20\,FPS,
and 3,000 at 60\,FPS; median duration is 14.38\,s (IQR 8.83--19.07\,s).
The 3DGS branch has 35,000 450-frame and 700 1,800-frame videos at 30\,FPS.
Long videos are the first 100 validated trajectory IDs per scene; the
short-video rerender manifest links each export to its original trajectory.

Path cost $J(\tau)$ is an unnormalized cumulative trajectory objective balancing translation distance, angular smoothness, and directional continuity.
Reconstruction adapters convert recorded positions, camera bases, FOV,
handedness, and units to fixed poses. Policy scene splits follow original environment identities. Reconstruction view partitions organize the capture views inside each environment.

\subsection{Training subset and action extraction}
\label{app:data-training}
Experiment-specific training subsets are drawn from this corpus inventory.
The 10/3/36 training/development/unseen scene split follows original environment identities, as specified in
Appendix~\ref{app:benchmark}.

\paragraph{Recorded time mapping and action extraction.}
RE9 conversion matches planned key poses to engine logs and estimates a
constant offset $\widehat\Delta$ from median inlier matches. Frame $f$ at
rate $F$ queries time $\tau_f=f/F+\widehat\Delta$; position, unwrapped yaw,
and pitch are interpolated on that time base. 3DGS conversion reads indexed
cameras at every second sample of 30\,FPS exports (15\,Hz). Local camera actions
$u_t=(\delta\mathbf p_t,\delta R_t)\in\mathrm{SE}(3)$ from consecutive camera-to-world poses
$(\mathbf p_t,R_t)$ are
\begin{equation}
 \delta\mathbf p_t=R_t^\top(\mathbf p_{t+1}-\mathbf p_t),\qquad
 \delta R_t=R_t^\top R_{t+1}.
 \label{eq:data-action-extraction}
\end{equation}
This defines the stored frame--pose synchronization used for action extraction.

\paragraph{Ten-dimensional action representation.}

The 10-D action vector encodes 3D local translation ($\delta\mathbf{p}_t\in\mathbb{R}^3$) and 6D continuous rotation formed from the first two columns of $\delta R_t$, plus one constant-zero reserved interface slot excluded by the mask $M$ in Eq.~\ref{eq:policy-flow}.
Frame indices and camera poses stay synchronized on the recorded time base above.

\subsection{Source processing and release scope}
\label{app:data-release}

SpatialVID-HQ~\citep{wang2025spatialvid} supplies a separate inventory of
347,111 retained Internet clips with estimated relative-scale poses, curated by sampling video frames to evaluate $s_1(I)$ scores, identifying smooth score-ascending camera trajectories, and cropping uniform-length segments;
these clips are outside the 96,700 constructed trajectories.
Retained/excluded manifests accompany the resource.

\paragraph{Release scope.}
The supplement provides anonymized human judgments and reproducible
summaries. Inventories, processing code, camera trajectories, scores, and
permitted media/assets are organized with source provenance for release.
Third-party access follows the applicable agreements; SpatialVID retains
its upstream CC-BY-NC-SA-4.0 terms.

\subsection{3DGS scene construction}
\label{app:data-infra}

\paragraph{Inventory and source identity.}

The 49-asset library stores calibration, coordinate transforms, allowed
volumes, source identities, and volume hashes (Table~\ref{tab:infra-inventory}).
The seven 3DGS demonstration collections map to Wukong assets in
Table~\ref{tab:gs-scene-map}; duration and reconstruction variants share
their original-scene identity for policy splitting.

\begin{table*}[t]
\centering\renewcommand{\arraystretch}{1.08}
\small
\caption{\textbf{The 49-scene 3DGS asset inventory} (scene-level splits follow original environment identities; Appendix~\ref{app:benchmark}).}
\label{tab:infra-inventory}
\begin{tabular}{@{}lrp{.68\textwidth}@{}}
\toprule
Source group & Assets & Composition \\
\midrule
Game & 12 & Black Myth: Wukong (8), RE9 (3), Kingdom Come: Deliverance II (1) \\
Interior & 15 & Fourteen interior assets and Playroom \\
Object-centric & 8 & Chair, drums, ficus, hotdog, lego, materials, mic, ship \\
Campus / aerial & 4 & Campus captures (3), drone capture (1) \\
Additional scenes & 10 & Babylon Lion, Christmas Stitch, Devils Tower, Everest Drive, Fongle Park, Lighthouse, Lockworks Tavern, Park Bench, Tesla car, Toucheng Temple \\
\midrule
Total & 49 & Each has a coordinate transform and an allowed-position volume \\
\bottomrule
\end{tabular}
\end{table*}

\begin{table}[!htb]
\centering\renewcommand{\arraystretch}{1.08}\small
\caption{\textbf{3DGS demonstration collections and scene identities} (5,000 short and 100 long trajectories per Wukong scene).}
\label{tab:gs-scene-map}
\resizebox{\linewidth}{!}{%
\begin{tabular}{@{}lll@{}}
\toprule
Collection & Captured environment & Asset identifier \\
\midrule
02 & Huaguoshan & \texttt{b1-huaguoshan} \\
03 & Pansidong & \texttt{b1-pansidong} \\
04 & Ziyunshan & \texttt{b1-ziyunshan} \\
05 & Xiaoxitian: Zhaojinghu & \texttt{b1-xiaoxitian-zhaojinghu} \\
06 & Pansiling: Zhujiadayuan & \texttt{b1-pansiling-zhujiadayuan} \\
07 & Huangfengling: Sihaliguo & \texttt{b1-huangfengling-sihaliguo} \\
08 & Xiaoxitian: Kuhaibeian & \texttt{b1-xiaoxitian-kuhaibeian} \\
\bottomrule
\end{tabular}%
}
\end{table}
The mapping follows the release and runtime asset catalog. RE9 batches
1.1--1.3 also share one original scene and the scene-1 reconstruction.
Policy scene splits follow these original-scene identities and are specified in Appendix~\ref{app:benchmark}.

\paragraph{Capture and reconstruction.}

Dense oblique captures supplement position-indexed images with overlapping
views at different heights and directions. Recorded poses, camera bases,
and FOV determine handedness and intrinsics before neighbor matching and
fixed-pose triangulation. OpenFlyScan~\citep{you2026openflyscanqualityguidedaerialreconstruction}
reconstructs these images; exported Gaussians retain native coordinates and
are rendered with EWA splatting.

\begin{figure*}[t]
 \centering
 \includegraphics[width=\textwidth]{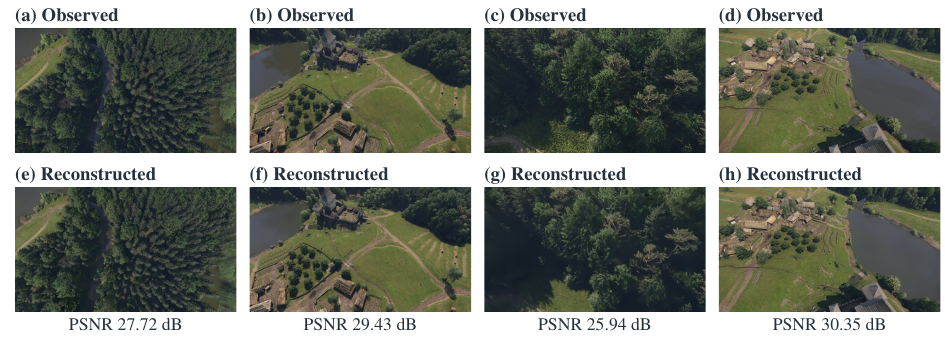}
 \caption{\textbf{Held-out KCD2 3DGS reconstruction comparison.} Top: ground-truth views withheld from photometric fitting; bottom: full-SH3 renderings.}
 \label{fig:infra-reconstruction}
\end{figure*}

\paragraph{Reconstruction diagnostics.}
\label{app:reconstruction-fidelity}
Recorded source/render comparisons quantify image reconstruction error for
KCD2 lake and RE9 scene 3.1 (Table~\ref{tab:reconstruction-fidelity}).
KCD2 uses 128 cameras sampled from 1,166 views excluded from photometric
fitting but included in triangulation and construction-time checks.
Measurements are computed on unmasked $1280\times720$ RGB images with
full SH3. RE9 values follow its archived 128-view evaluation report.
Figure~\ref{fig:infra-reconstruction} shows fixed KCD2 examples. These held-out view diagnostics confirm high photometric fidelity across novel viewpoints in our reconstructed 3DGS environments.
\begin{table}[!htb]
\centering\renewcommand{\arraystretch}{1.08}\small
\caption{\textbf{Held-out 3DGS reconstruction PSNR (dB) at recorded capture cameras.}}
\label{tab:reconstruction-fidelity}
\resizebox{\linewidth}{!}{%
\begin{tabular}{@{}lrrrr@{}}
\toprule
Scene & Views & Mean & Median & 10th percentile \\
\midrule
KCD2 lake & 128 & 27.48 & 28.33 & 24.18 \\
RE9 scene 3.1 & 128 & 33.39 & 33.93 & 30.27 \\
\bottomrule
\end{tabular}%
}
\end{table}

\section{Full-corpus data validation}
\label{app:data-audit}

\subsection{Complete-inventory results}
\label{sec:data-validation}

The evaluation scores all 44,264,749 decoded frames across all 96,700 retained trajectories with all three aesthetic models ($s_1$, $s_{2.5}$, and $s_{\mathrm{AM}}$) to verify the effectiveness of the $s_1$-collected data beyond the fast collection scorer. Internet clips are inventoried separately from these game and 3DGS trajectories.
Every retained trajectory remains in the evaluation after replay.
For each video, we compute endpoint gain
$\Delta_\tau=s_{\tau,T_\tau-1}-s_{\tau,0}$ and the time--score Spearman
correlation $\rho_\tau$, weighting trajectories equally, where $s_{\tau,t}=s_{\mathrm{ens}}(I_{\tau,t})$ is the ensemble score of trajectory $\tau$ at frame $t$.
Table~\ref{tab:data-full-audit} summarizes construction and replay
statistics; Figure~\ref{fig:data-full-corpus}c shows the corpus-level score progression and endpoint gains.

\begin{table}[!ht]
 \centering\renewcommand{\arraystretch}{1.08}
 \small
 \caption{\textbf{Full-corpus trajectory construction and aesthetic progression statistics} (equal per-trajectory weighting).}
 \label{tab:data-full-audit}
 \resizebox{\linewidth}{!}{%
 \begin{tabular}{@{}lrr@{}}
 \toprule
 Quantity & Game & 3DGS \\
 \midrule
 Retained trajectories & 61,000 & 35,700 \\
 Decoded and scored frames & 27,254,749 & 17,010,000 \\
 Original key-view occurrences & 485,904 & 149,956 \\
 Original key views per trajectory & 5--13 & 4--6 \\
 Adjacent key-view transitions & 424,904 & 114,256 \\
 Strictly increasing key-view sequences & 100.00\% & 100.00\% \\
 Median minimum key-view increment & 0.204 & 0.380 \\
 \midrule
 Videos with positive endpoint gain & 99.74\% & 100.00\% \\
 Median aesthetic-score gain & $+2.521$ & $+2.437$ \\
 Videos with positive time--score correlation & 99.69\% & 99.96\% \\
 Median time--score Spearman correlation & 0.916 & 0.849 \\
 Positive adjacent-frame comparisons & 51.41\% & 51.64\% \\
 Positive comparisons at 10\% duration & 71.08\% & 69.98\% \\
 Positive comparisons at 25\% duration & 87.62\% & 80.11\% \\
 Positive comparisons at 50\% duration & 96.80\% & 92.87\% \\
 \bottomrule
 \end{tabular}%
 }
\end{table}

\paragraph{Component agreement.}

Of 35,700 component-audit videos, 35,584 (99.68\%) improve at the endpoints
under every predictor, and 34,307 (96.10\%) have positive time--score
correlations under all three (Table~\ref{tab:data-scorer-components}).
Strong agreement across all three distinct backbones confirms consistent
trajectory-level aesthetic gains, complemented by our independent human
preference study in Appendix~\ref{app:data-human}.

\begin{table}[!ht]
 \centering\renewcommand{\arraystretch}{1.08}
 \small
 \caption{\textbf{Per-predictor aesthetic progression on 35,700 trajectories.} All three scoring backbones consistently agree on trajectory-level gains.}
 \label{tab:data-scorer-components}
 \resizebox{\linewidth}{!}{%
 \begin{tabular}{@{}lrrrr@{}}
 \toprule
 Measure & Positive gain & Median $\Delta_\tau$
        & Positive $\rho_\tau$ & Median $\rho_\tau$ \\
 \midrule
 $s_1$ & 100.00\% & $+1.901$ & 99.20\% & 0.715 \\
 $s_{2.5}$ & 99.68\% & $+1.297$ & 97.04\% & 0.703 \\
 $s_{\mathrm{AM}}$ & 100.00\% & $+4.091$ & 99.76\% & 0.803 \\
 Ensemble $s_{\mathrm{ens}}$ & 100.00\% & $+2.437$ & 99.96\% & 0.849 \\
 \bottomrule
 \end{tabular}%
 }
\end{table}

\subsection{Replay statistics and plotting}
Statistics include all retained frames; Spearman correlation uses frame
order with averaged tied ranks. For a lag of $k$ frames,
\begin{equation}
 p_\tau(k)=\frac{1}{T_\tau-k}\sum_{t=0}^{T_\tau-k-1}
     \1[s_{\tau,t+k}>s_{\tau,t}].
 \label{eq:data-lagged-comparison}
\end{equation}
Here $s_{\tau,t}=s_{\mathrm{ens}}(I_{\tau,t})$ is the ensemble score at frame $t$.
Rates at 10\%, 25\%, and 50\% of the frame-index span weight trajectories
equally and retain ties/negative comparisons in the denominator.
Reported statistics use the original frame indices.
Plots resample relative scores to 101 normalized times and show median and
quartiles; endpoint histograms use 48 density-normalized bins.

\subsection{Many-to-few aesthetic basin convergence}
\label{app:data-coverage}

We count distinct planned poses within each collection coordinate system,
using position and yaw/pitch rounded to $10^{-6}$ native units and radians,
with periodic yaw canonicalized. The 15-second and 60-second 3DGS subsets
are merged by their recorded scene identity; game batches remain separate
collection groups within each source environment.
If $K$ planning-view occurrences use $U$ distinct poses, pose reuse is
$1-U/K$. Directed-edge reuse uses the same ratio over consecutive ordered
pose pairs and measures how often a planned transition recurs across trajectories.

\begin{table*}[t]
\centering\renewcommand{\arraystretch}{1.08}\small
\caption{\textbf{Many-to-few aesthetic basin convergence across planning records.} Diverse starts funnel into top local aesthetic maxima (\emph{Top end}: largest endpoint share; \emph{Reuse}: repeat fraction across poses or edges).}
\label{tab:data-planning-coverage}
\begin{tabular}{@{}lrrrrrrr@{}}
\toprule
Collection & Traj. & Poses & Starts & Ends & Top end & Pose reuse & Edge reuse \\
\midrule
B1 Heifengdong & 3,000 & 1,235 & 447 & 1 & 100.00\% & 91.8\% & 70.5\% \\
RE9 batch 1.1 & 4,000 & 1,870 & 1,255 & 1 & 100.00\% & 94.8\% & 87.6\% \\
RE9 batch 1.2 & 10,000 & 3,861 & 2,403 & 4 & 25.00\% & 95.1\% & 86.3\% \\
RE9 batch 1.3 & 3,000 & 90 & 2 & 1 & 100.00\% & 99.7\% & 98.3\% \\
RE9 batch 2 & 13,000 & 3,816 & 1,520 & 1 & 100.00\% & 96.1\% & 86.1\% \\
RE9 batch 3.1 & 15,000 & 3,885 & 1,635 & 1 & 100.00\% & 96.6\% & 88.7\% \\
RE9 batch 3.2 & 13,000 & 2,415 & 893 & 1 & 100.00\% & 97.9\% & 93.8\% \\
3DGS scene 02 & 5,100 & 3,833 & 168 & 2 & 98.04\% & 81.7\% & 27.1\% \\
3DGS scene 03 & 5,100 & 3,002 & 158 & 1 & 100.00\% & 86.0\% & 30.3\% \\
3DGS scene 04 & 5,100 & 3,110 & 174 & 2 & 98.04\% & 85.4\% & 29.8\% \\
3DGS scene 05 & 5,100 & 3,341 & 163 & 1 & 100.00\% & 84.3\% & 28.6\% \\
3DGS scene 06 & 5,100 & 3,437 & 181 & 1 & 100.00\% & 84.2\% & 26.9\% \\
3DGS scene 07 & 5,100 & 3,102 & 115 & 1 & 100.00\% & 85.4\% & 28.3\% \\
3DGS scene 08 & 5,100 & 4,247 & 133 & 1 & 100.00\% & 80.6\% & 21.4\% \\
\bottomrule
\end{tabular}
\end{table*}

All 96,700 ordered key-pose sequences are distinct, forming dense many-to-few convergence fields around canonical 3D compositions (Table~\ref{tab:data-planning-coverage}).
Within each environment, hundreds to thousands of diverse starting viewpoints and intermediate camera paths funnel consistently toward top local aesthetic maxima, teaching the policy to recover reliably from arbitrary off-axis framings while the 347,111 Internet video clips supply broad open-world visual priors.

\subsection{Replay scores at fixed elapsed times}
\label{app:data-fixed-time}

To characterize aesthetic change at fixed elapsed times, we evaluate every valid
frame pair at $\Delta t\in\{0.5,1,2\}$ seconds, using each video's frame
rate. For each trajectory we compute the positive fraction,
median, and fifth percentile of the score differences, with ties kept in the denominator.
Table~\ref{tab:data-fixed-time}
averages positive fractions equally across trajectories and reports the
median of those trajectory-specific quantiles.

\begin{table}[!htb]
\centering\renewcommand{\arraystretch}{1.08}\small
\caption{\textbf{Aesthetic score progression at fixed elapsed intervals across 44.3M frames.} Longer horizons yield more reliable positive gains.}
\label{tab:data-fixed-time}
\resizebox{\linewidth}{!}{%
\begin{tabular}{@{}llrrrr@{}}
\toprule
Source & Interval & Positive & Median gain & 5th-percentile gain & Worst decline \\
\midrule
Game & 0.5 s & 60.62\% & 0.077 & -0.475 & 1.037 \\
Game & 1 s & 67.91\% & 0.176 & -0.498 & 1.051 \\
Game & 2 s & 78.54\% & 0.381 & -0.384 & 0.916 \\
3DGS & 0.5 s & 61.61\% & 0.071 & -0.339 & 0.713 \\
3DGS & 1 s & 66.50\% & 0.139 & -0.392 & 0.783 \\
3DGS & 2 s & 72.49\% & 0.276 & -0.421 & 0.801 \\
\bottomrule
\end{tabular}%
}
\end{table}

Longer elapsed intervals raise the rate of positive score change
(Table~\ref{tab:data-fixed-time}).

\begin{figure*}[t]
 \centering
 \includegraphics[width=\textwidth]{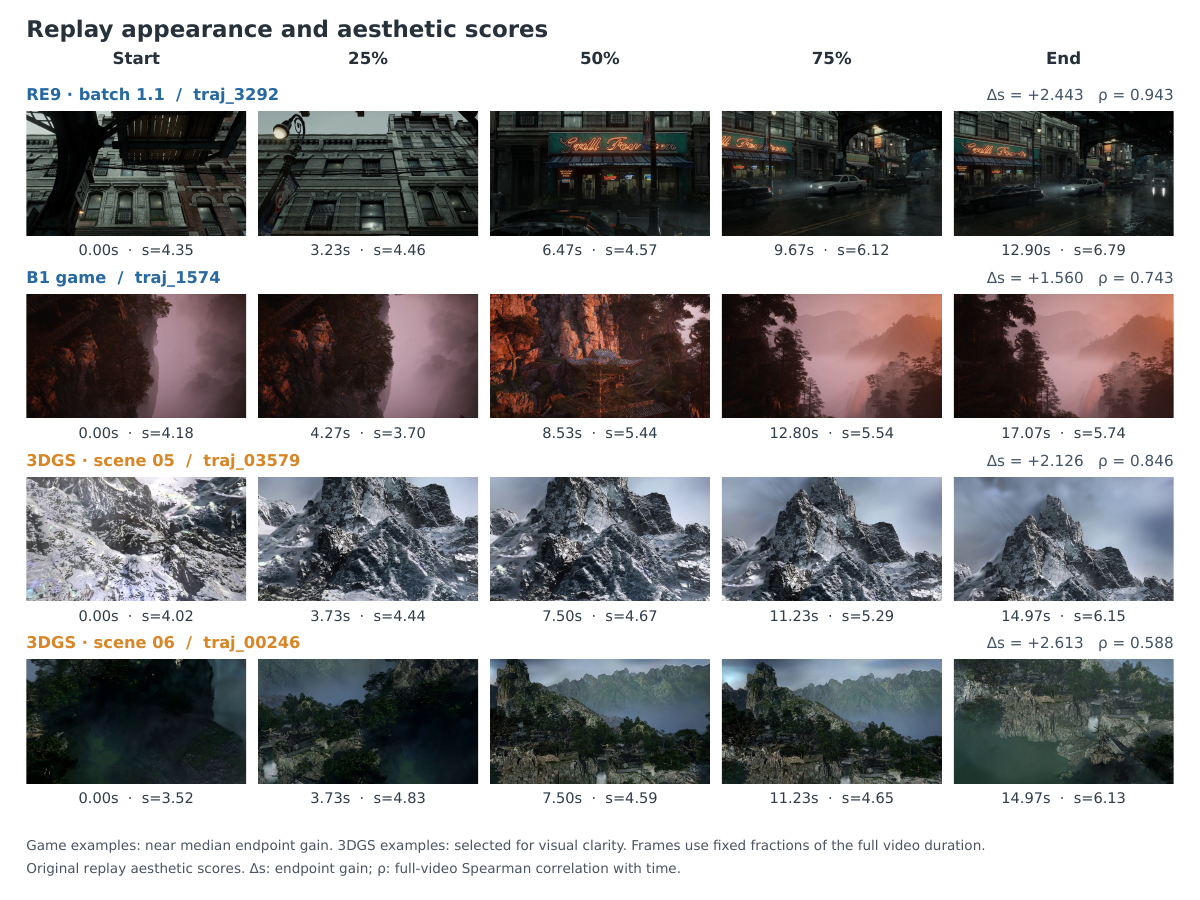}
 \caption{\textbf{Representative replayed trajectories at 0/25/50/75/100\% progress} ($\Delta s_{\mathrm{ens}}$: endpoint ensemble gain; $\rho$: time--score Spearman correlation).}
 \label{fig:ezcam_visual_pose_gallery}
\end{figure*}

\paragraph{32-step action windows.}
Each exported window contains 33 views and therefore 32 consecutive action intervals.
The analysis uses 13,930 game windows and all 35,700 3DGS windows.
Their elapsed duration is $2.13\,\mathrm{s}$: 2.10--2.15\,s for games and 2.133\,s for 3DGS, with frame-index rounding accounting for the game variation.
Table~\ref{tab:data-action-windows} weights these windows equally.
Adjacent steps are positive on 52.80\% of game steps and 51.74\% of 3DGS steps (${\sim}52\%$), whereas the same windows have positive 32-step chunk gains on 85.87\% of game windows and 69.62\% of 3DGS windows.
This transition from noisy single-step differences to reliable multi-step chunk gains validates Insight~1 (representative replayed progressions are shown in Figure~\ref{fig:ezcam_visual_pose_gallery}).

\paragraph{Action--score coupling.}
Across 32-step windows, unconditioned scalar motion features (stepwise translation distance, rotation rate, and path straightness) exhibit near-zero linear correlation with state-residualized aesthetic gain ($|r|<0.08$), confirming that raw movement magnitude alone does not drive aesthetic improvement.
In contrast, clip-disjoint (60/20/20) action-conditional MLP probes show that adding 32-step 6-DoF action encodings to current score context substantially boosts held-out future aesthetic-gain explained variance by $\Delta R^2 = +0.240\pm 0.023$ on game-engine trajectories ($n=1{,}500$ balanced RE9 windows; $+0.261\pm 0.038$ over within-split shuffled actions) and $\Delta R^2 = +0.293\pm 0.004$ on 3DGS trajectories ($n=5{,}040$ Wukong windows; $+0.252$ over frozen CLIP ViT-L/14 visual features and $+0.318\pm 0.007$ over shuffled actions), demonstrating strong 6-DoF action--aesthetic predictability across metrically calibrated 3D scenes.

\begin{table*}[t]
\centering\renewcommand{\arraystretch}{1.08}\small
\caption{\textbf{Single-step vs.\ 32-step ($2.13\,\mathrm{s}$) window score changes.} Multi-step chunks convert noisy adjacent steps (\emph{Step+}) into reliable positive gains (\emph{Chunk+}), supporting Insight 1.}
\label{tab:data-action-windows}\label{tab:data-main-windows}
\begin{tabular}{@{}lrrrrrr@{}}
\toprule
Source & Windows & Step+ & Chunk+ & Gain [5th, 95th] & Below start & Drawdown \\
\midrule
Game & 13,930 & 52.80\% & 85.87\% & 0.415 [$-0.219$, 1.028] & 29.56\% & 0.669 \\
3DGS & 35,700 & 51.74\% & 69.62\% & 0.207 [$-0.490$, 1.012] & 42.54\% & 0.502 \\
\bottomrule
\end{tabular}
\end{table*}

\section{Human preference comparison}
\label{app:data-human}

\paragraph{Participants and inclusion.}

All 37 recruited participants (fine-arts students and members of the public)
completed 60 pairs, yielding 2,220 judgments. Every participant and every response enters the analysis.

\paragraph{Pair construction and presentation.}

The 60 distinct trajectories cover 29 collection subsets: 21 game and 39
3DGS pairs, including 42 from the main corpus and 18 from additional
collections. Sampling covers these collection subsets.
Each pair compares the trajectory's initial frame with a frame at one of the times 0, 0.5, \ldots, 10\,s.
Both images use the same score, and all 120 values match the frozen manifest.

Participants see images without scores or frame roles and choose A, B,
similar, or cannot judge. The fixed material mapping places the two frames
on each side in 30 pairs. The client randomizes pair order and presentation
side, then restores fixed labels for response recording.
The comparison evaluates agreement on still-image aesthetics.

\paragraph{Outcome definition and complete accounting.}

\begin{figure*}[t]
\centering
\includegraphics[width=\textwidth]{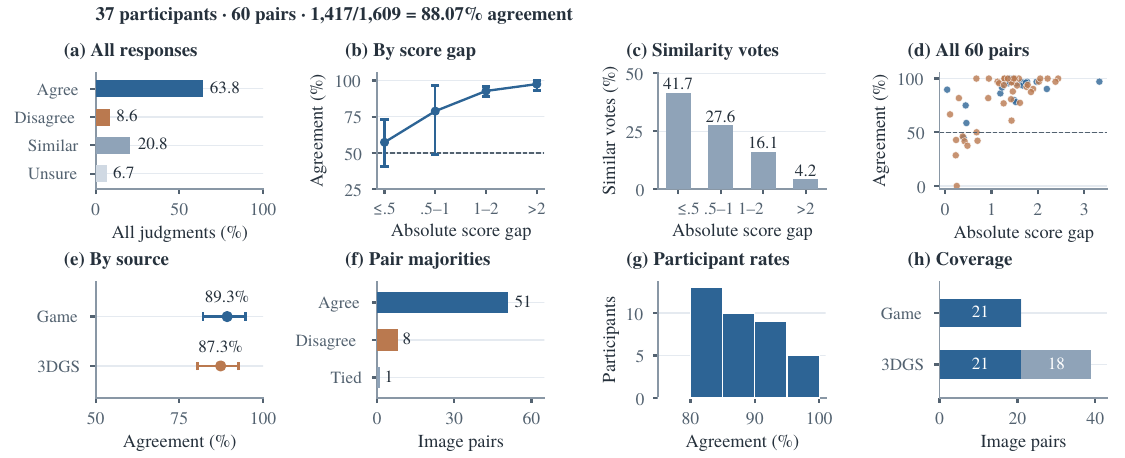}
\caption{\textbf{Blinded human preference study (37 participants, 60 pairs; error bars: two-way 95\% bootstrap CI).} Human preferences strongly align with $s_{\mathrm{ens}}(I)$, especially as score gaps widen.}
\label{fig:app-human-panels}
\end{figure*}

For pair $i$, let $d_i=s_{\mathrm{ens}}(I_i^{\mathrm{high}})-s_{\mathrm{ens}}(I_i^{\mathrm{low}})$.
All 60 pairs have nonzero score differences $d_i>0$.
Agreement is a directional choice of the higher-$s_{\mathrm{ens}}(I)$ view in the pair.
Directional agreement is $1{,}417/1{,}609=88.07\%$ across the 1,609 directional preferences, with the 462 similar and 149 cannot-judge responses documented in full in Table~\ref{tab:data-human-accounting}.
Figure~\ref{fig:app-human-panels} plots every group in that accounting.

\begin{table}[!ht]
 \centering\renewcommand{\arraystretch}{1.08}
 \small
 \caption{\textbf{Human preference response breakdown across all 37 participants} (2,220 judgments; directional agreement excludes \emph{Similar} and \emph{Cannot judge}).}
 \label{tab:data-human-accounting}
 \resizebox{\linewidth}{!}{%
 \begin{tabular}{@{}lrrrrr@{}}
 \toprule
 Group & Pairs & Agree/clear & Agreement & Similar & Cannot judge \\
 \midrule
 All pairs & 60 & 1417/1609 & 88.07\% & 462 & 149 \\
 \midrule
 $\leq 0.5$ & 12 & 117/204 & 57.35\% & 185 & 55 \\
 $(0.5,1]$ & 5 & 82/104 & 78.85\% & 51 & 30 \\
 $(1,2]$ & 36 & 986/1063 & 92.76\% & 215 & 54 \\
 $>2$ & 7 & 232/238 & 97.48\% & 11 & 10 \\
 \midrule
 Game & 21 & 541/606 & 89.27\% & 152 & 19 \\
 3DGS & 39 & 876/1003 & 87.34\% & 310 & 130 \\
 \midrule
 Main corpus & 42 & 1071/1191 & 89.92\% & 295 & 68 \\
 Additional collections & 18 & 346/418 & 82.78\% & 167 & 81 \\
 \bottomrule
 \end{tabular}%
 }
\end{table}

\paragraph{Uncertainty estimates.}

The 95\% interval, 83.2--92.2\%, uses 20,000 two-way percentile bootstrap
replicates (seed 20260920). Each independently resamples 60 pairs and 37
participants, then recomputes agreement on the crossed sample. For grouped
estimates, pairs are resampled within each group.
Participant majorities concur with $s_{\mathrm{ens}}(I)$ on 51 of the 59 non-tied pairs (86.4\%).

\paragraph{Interpretation of score differences.}

While subtle sub-0.5 score differences naturally elicit frequent \emph{Similar} judgments near human perceptual thresholds, directional agreement rises sharply from 78.85\% at moderate gaps $(0.5,1]$ to 92.76\%--97.48\% across substantial compositional gains ($d_i>1$; Table~\ref{tab:data-human-accounting}).
Main-corpus pairs give 89.92\% agreement (95\% interval 84.9--94.0\%);
additional pairs give 82.78\% (70.1--92.0\%), using 20,000 two-way replicates
and seed 20260922.
These frozen-pair results measure how consistently participants select the higher-$s_{\mathrm{ens}}(I)$ view.

\subsection{Component diagnostics on the human study}
\label{app:human-components}

Component comparisons use the same 120 images and 1,609 directional
judgments, with frozen score weights. Tied component scores count as non-agreement:
$s_{2.5}$ ties on two pairs; the other predictors have none. Paired
two-way bootstrap differences share each resampled participant--pair matrix
across predictors (20,000 replicates; seed 20260922).

\begin{table}[!htb]
\centering\renewcommand{\arraystretch}{1.08}\small
\caption{\textbf{Human agreement of individual aesthetic predictors vs.\ ensemble $s_{\mathrm{ens}}(I)$} (differences in percentage points with paired 95\% bootstrap CI).}
\label{tab:human-components}
\resizebox{\linewidth}{!}{%
\begin{tabular}{@{}lrrl@{}}
\toprule
Component / score & Agree/clear & Agreement & $s_{\mathrm{ens}}$ minus component (95\% interval) \\
\midrule
$s_1$ (LAION V1) & 1414/1609 & 87.88\% & +0.19 [-0.19, 1.05] \\
$s_{2.5}$ (V2.5) & 1308/1609 & 81.29\% & +6.77 [0.81, 14.13] \\
$s_{\mathrm{AM}}$ (ArtiMuse) & 1392/1609 & 86.51\% & +1.55 [-1.95, 5.95] \\
Ensemble $s_{\mathrm{ens}}$ & 1417/1609 & 88.07\% & --- \\
\bottomrule
\end{tabular}%
}
\end{table}

The ensemble score $s_{\mathrm{ens}}(I)$ achieves the highest human agreement ($88.07\%$) among all evaluated predictors (Table~\ref{tab:human-components}), while mitigating single-backbone variance across synthetic and real domains.

\subsection{Double-blind human evaluation of policy outputs (\texorpdfstring{$N=106$}{N=106})}
\label{app:policy-human}
\begin{figure*}[t]
\centering
\includegraphics[width=0.92\textwidth]{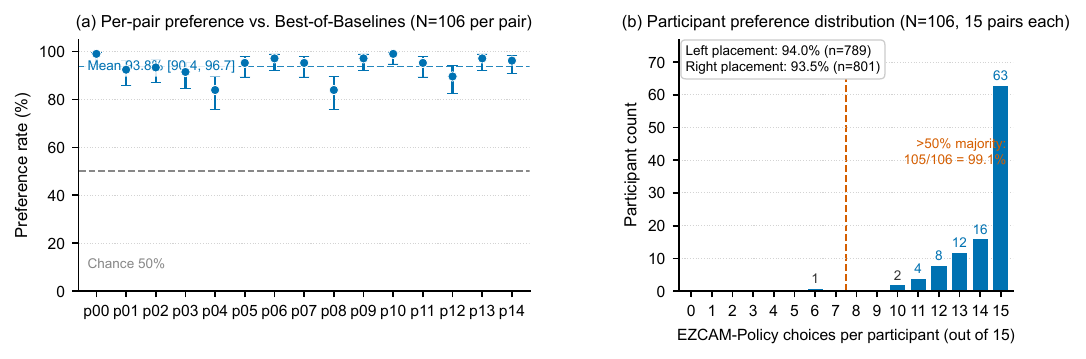}
\caption{\textbf{Double-blind 2AFC human preference study on policy outputs ($N=106$ participants, 15 pairs, 1,590 judgments) against per-scene Best-of-Baselines.}
\emph{(a)} Per-pair EZCAM-Policy win rate with 95\% Wilson CIs ($15/15$ pairs $>83.9\%$).
\emph{(b)} Participant preference distribution ($105/106$ participants above 50\% chance) and left/right placement balance.}
\label{fig:app-policy-human-106}
\end{figure*}

To directly validate closed-loop policy outputs with human judges independently of automated aesthetic predictors, we conducted a separate double-blind two-alternative forced-choice (2AFC) study with \textbf{106 independent participants} across 15 representative benchmark scene pairs (1,590 total judgments; Figure~\ref{fig:app-policy-human-106} and Table~\ref{tab:policy-human-106}).

\paragraph{Adversarial Best-of-Baselines protocol and randomization.}
In each sampled scene and shared starting pose under the PF41 protocol, we pair the photograph captured by \textbf{EZCAM-Policy} against the photograph captured by the \textbf{strongest competing baseline on that scene} (\emph{Per-Scene Best-of-Baselines}), rather than comparing against a single fixed baseline that may degrade on specific scene domains.
All 30 stimulus images are anonymized with irreversible 16-character hexadecimal hashes, and left/right display positions are randomized independently on every trial (789 left vs.\ 801 right placements for EZCAM-Policy across all 1,590 trials). Every recruited participant completed all 15 pairs with zero duplicate or invalid records.

\paragraph{Statistical results across participants, pairs, and crossed random effects.}
Participants selected \textbf{EZCAM-Policy} in \textbf{1,491 of 1,590} trials (\textbf{93.77\%} overall preference win rate vs.\ 6.23\% for Best-of-Baselines):
\emph{(i) Crossed uncertainty:} Accounting simultaneously for participant and stimulus-pair variance, a crossed random-effects model (Satterthwaite $\mathrm{df}=26.77$, $\mathrm{SE}=1.47\%$) yields a 95\% CI of \textbf{90.76--96.79\%}, matching a 20,000-replicate two-way percentile bootstrap resampling both 106 participants and 15 pairs (\textbf{90.38--96.67\%}).
\emph{(ii) Pair-level consistency:} \textbf{EZCAM-Policy} won a clear majority on \textbf{all 15 of 15 pairs} (mean $93.77\%$, SD $4.78\%$, 95\% CI $91.13\text{--}96.42\%$, two-sided $t_{14}=35.46$, $p=4.13\times 10^{-15}$; exact sign test $p=6.1\times 10^{-5}$), ranging from $83.96\%$ ($89/106$ on \texttt{p004} and \texttt{p008}) to $99.06\%$ ($105/106$ on \texttt{p000} and \texttt{p010}; median $95.28\%$).
\emph{(iii) Participant-level consistency and position balance:} \textbf{105 of 106 participants} ($99.06\%$) preferred EZCAM-Policy on $>50\%$ of pairs (mean $14.07/15$, median $15/15$; 63 participants chose EZCAM-Policy on all $15/15$ pairs, 16 on $14/15$, 12 on $13/15$, eight on $12/15$, four on $11/15$, two on $10/15$, and one on $6/15$; two-sided $t_{105}=44.85$, Wilcoxon signed-rank $p<10^{-16}$). Preference rates were invariant to screen side ($94.04\%$ when placed left vs.\ $93.51\%$ when placed right; total left/right clicks: $794/796$).

\begin{table}[!htb]
\centering\renewcommand{\arraystretch}{1.08}\small
\setlength{\tabcolsep}{5.5pt}
\caption{\textbf{Per-pair breakdown of the $N=106$ double-blind policy-output human study against per-scene Best-of-Baselines} (1,590 total votes; L/R place: EZCAM left/right count; L/R click: participant left/right choices).}
\label{tab:policy-human-106}
\resizebox{\linewidth}{!}{%
\begin{tabular}{@{}lrrrr@{}}\toprule
Pair ID & EZCAM / Votes & Win Rate (\%) & L / R Place & L / R Click \\\midrule
\texttt{p000} & 105 / 106 & 99.06 & 52 / 54 & 53 / 53 \\
\texttt{p001} & 98 / 106 & 92.45 & 45 / 61 & 49 / 57 \\
\texttt{p002} & 99 / 106 & 93.40 & 63 / 43 & 62 / 44 \\
\texttt{p003} & 97 / 106 & 91.51 & 48 / 58 & 45 / 61 \\
\texttt{p004} & 89 / 106 & 83.96 & 62 / 44 & 61 / 45 \\
\texttt{p005} & 101 / 106 & 95.28 & 59 / 47 & 56 / 50 \\
\texttt{p006} & 103 / 106 & 97.17 & 50 / 56 & 51 / 55 \\
\texttt{p007} & 101 / 106 & 95.28 & 52 / 54 & 53 / 53 \\
\texttt{p008} & 89 / 106 & 83.96 & 45 / 61 & 46 / 60 \\
\texttt{p009} & 103 / 106 & 97.17 & 46 / 60 & 47 / 59 \\
\texttt{p010} & 105 / 106 & 99.06 & 45 / 61 & 46 / 60 \\
\texttt{p011} & 101 / 106 & 95.28 & 53 / 53 & 54 / 52 \\
\texttt{p012} & 95 / 106 & 89.62 & 61 / 45 & 58 / 48 \\
\texttt{p013} & 103 / 106 & 97.17 & 48 / 58 & 51 / 55 \\
\texttt{p014} & 102 / 106 & 96.23 & 60 / 46 & 62 / 44 \\\midrule
\textbf{Total / Mean} & \textbf{1,491 / 1,590} & \textbf{93.77} & \textbf{789 / 801} & \textbf{794 / 796} \\\bottomrule
\end{tabular}%
}
\end{table}

\section{Infrastructure details}
\label{app:infrastructure}
\subsection{Coordinate and camera conventions}
\label{app:infra-coordinates}
A scene rotation $Q_e\in\mathrm{SO}(3)$ maps native up to canonical $+Z$:
\begin{equation}
 R_c=Q_eR_n,\qquad \mathbf p_c=Q_e\mathbf p_n.
 \label{eq:infra-coordinate-map}
\end{equation}
The transformation preserves native origin and scale. Camera rotation
columns encode image-right, image-down, and optical-forward. EWA rendering
uses 8-bit sRGB and
\begin{equation}
 \begin{aligned}
 K&=\begin{bmatrix}320&0&320\\0&320&272\\0&0&1\end{bmatrix},\\
 (W,H)&=(640,544),\qquad \mathrm{HFOV}=90^\circ.
 \end{aligned}
 \label{eq:infra-camera}
\end{equation}

\subsection{Allowed domains and continuous execution}
\label{app:infra-geometry}
Cell centers $\mathbf c_j$ and scene width $h_e$ define
\begin{equation}
 \mathcal V_e=\bigcup_j C_j,\qquad
 C_j=\{\mathbf p:\|\mathbf p-\mathbf c_j\|_\infty\leq h_e/2\}.
 \label{eq:infra-volume}
\end{equation}
The cell construction uses collision grids, occupancy proxies, reviewed
viewpoint constraints, interior free space, or object-centric shells,
depending on the asset family. Cell sets are frozen before evaluation.
At runtime, the camera is a point. Slab intersections of
$\mathbf p(t)=\mathbf p_0+t(\mathbf p_1-\mathbf p_0)$ with each cell give
intervals $J_j\subseteq[0,1]$. Their union must cover the whole segment;
contact with a declared obstacle also rejects the action. The full pose is
preserved on rejection. The implementation uses absolute tolerance $10^{-9}$
in native coordinate or segment-parameter units.
Coordinate transforms $Q_e$ standardize orientation ($+Z$ up) while preserving each scene's native metric scale.

\subsection{Inputs, actions, and accounting}
\label{app:infra-interface}\label{app:infra-actions}\label{app:infra-accounting}
The policy receives its declared RGB, executed-action history, and internal LAION V1 ($s_1$) score
features. Pose, intrinsics, and allowed cells belong to the environment
record and executor. EZCAM-Policy's normalized action decoding is specified in
Section~\ref{sec:method}. Primary closed-loop benchmark runs evaluate $s_1(I)$ on observations already charged to the episode, and deterministic replay over all $194{,}070$ active unique $(e, \text{SHA-256})$ poses ($206{,}273$ cached frame evaluations across three shard groups; zero pixel-hash mismatches) evaluates held-out $s_{2.5}$, $s_{\mathrm{AM}}$, $s_{\mathrm{indep}}$, and $s_{\mathrm{ens}}$ (Appendix~\ref{app:heldout-shutter}).
Every future render is a charged observation.
Reset and each request produce a charged observation, including stationary
and rejected motion. Requested/executed actions, resulting poses, observations,
and action sources are retained in the episode record. A visited image is recorded at the pose where it was observed.

\subsection{Observation preprocessing}
\label{app:infra-preprocess}
Rendered $640\times544$ sRGB observations are resized to $224\times224$ and normalized with standard OpenCLIP image statistics before entering the visual encoder and frozen LAION V1 ($s_1$) scoring heads. Baseline adapters apply the native preprocessing specified by their respective backbones.

\subsection{Implementation checks}
\label{app:infra-verification}
Table~\ref{tab:infra-verification-summary} summarizes coordinate, execution,
and replay checks. The development pilot uses seven requests/eight
observations; final policy evaluation uses 40/41. The segment cases test
geometric execution validity.

\begin{table*}[t]
\centering\renewcommand{\arraystretch}{1.08}\small
\caption{\textbf{Closed-loop testbed verification and deterministic replay checks.}}
\label{tab:infra-verification-summary}
\begin{tabular}{@{}p{.28\textwidth}p{.26\textwidth}p{.40\textwidth}@{}}\toprule
Check & Population & Outcome\\\midrule
Spatial records & 49 assets & Metadata checks recorded\\
Reference cases & 32 cases & Expected outcomes match\\
Pilot transitions & 1,008 requests & 876 moves, 126 stationary, 6 rejected\\
Pilot RGB records & 1,152 observations & 894 distinct pixel arrays\\
Control replay & 15 cases / 30 images & Pose and image hashes match\\
Pose rerendering & 72 episodes / 576 images & Same-service hashes match\\
Long-segment geometry & 192 requests & 84 covered; 108 leave domain\\\bottomrule
\end{tabular}
\end{table*}

\subsection{Render-call cost and image coverage}
\label{app:infra-cost-views}
For 192 archived pilot observations per scene, client-side median/95th-percentile
render-call times are 143/178 ms (Wukong), 140/177 ms (KCD2), and 198/336 ms
(campus). They include transport and queueing on a shared service. All 576
images pass dimension and basic RGB checks.
Charging all observations, including boundary and sparse views, strictly enforces the closed-loop interaction budget without free previews.

\section{Evaluation protocol and controlled configurations}
\label{app:benchmark}
\label{app:completion}
PF41 evaluates six starts on each of 49 assets with 40 requests and
41 observations, without candidate-view previews.

\subsection{Shared evaluation settings}
All configurations use the same 49 assets, six prescribed starts per asset,
renderer, aesthetic scorer, and 40-request/41-observation budget. Starts are selected
from allowed positions and published cameras without aesthetic-score ranking.
Training scenes are s02--s08 and s79--s81; development scenes are s01, s33,
and s85. Asset aliases preserve the original environment identity used by this scene split.

Learned policies and VLN adapters use a per-step translation cap of $0.5$, a path budget of $20$ in native scene units, and a rotation cap of $15^\circ$, alongside two zero-inference motion heuristics (Appendix~\ref{app:motion-baselines}). Every configuration charges candidate-view renders to the interaction budget. Scorer, asset, and start identities are provided
in the supplementary manifests.

\subsection{Recorded outputs and inclusion rule}
\label{app:output-records}
Each complete record contains 40 requests and 41 scores. Selection returns
the earliest maximum, including the initial frame. The release includes
the shared starts, selected images, poses, and action traces.

Deterministic episode manifests log the executed trajectory, camera poses, and observation hashes for each prescribed scene and start configuration.

\subsection{Trained-model comparison and capture-pose independence}
\label{app:policy-comparison}
\begin{figure*}[t]
\centering
\includegraphics[width=\textwidth]{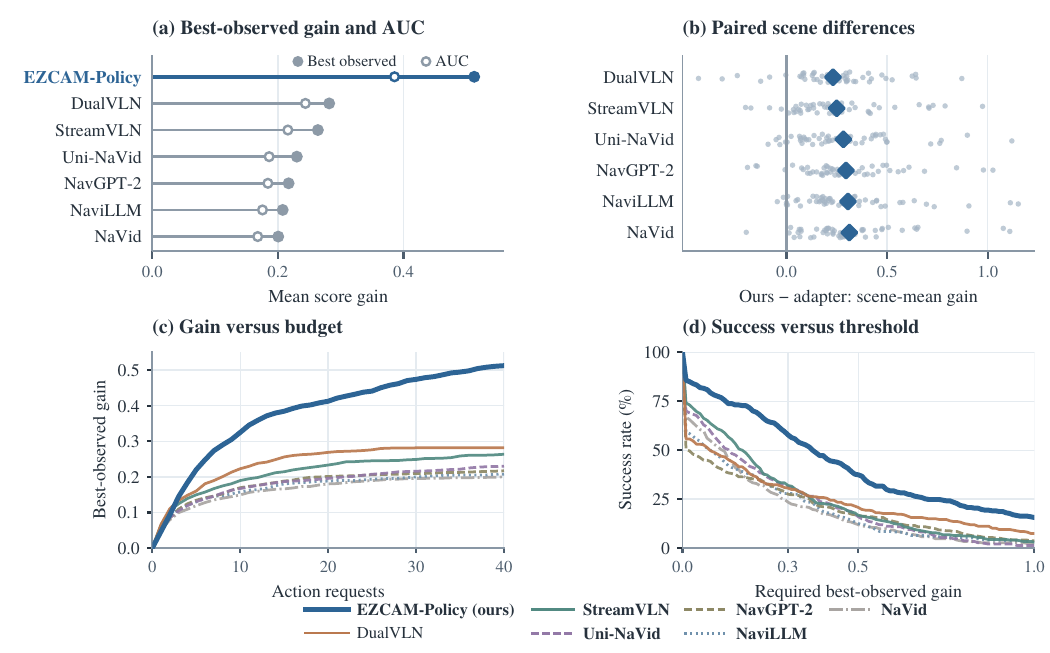}
\caption{\textbf{Detailed trained-model comparison across all 49 scenes ($n=294$, $T=40$).} \textbf{EZCAM-Policy} leads across scenes, budgets, and gain thresholds.}
\label{fig:policy-full49}
\label{fig:policy-budget}
\end{figure*}

Figure~\ref{fig:policy-full49} expands the 49-scene comparison in Table~\ref{tab:policy-main}.
EZCAM-Policy leads on best-observed gain and AUC, wins 43 of 49 scenes against DualVLN, reaches 0.3249 at 10 requests and 0.4126 at 20, and its success rate stays at or above every trained baseline at every required-gain threshold.
To verify that gains reflect genuine 3D composition at novel viewpoints rather than memorizing capture cameras or exploiting 3DGS floaters, we measure proximity between executed camera poses and dense capture-camera pools across all $N=11{,}760$ transitions on the 49 scenes: stepwise aesthetic gain $\Delta s_1$ exhibits near-zero correlation with normalized capture-camera distance (Pearson $r=+0.0085$, $p=0.35$; near vs.\ far distance bins: $-0.0090$ vs.\ $-0.0093$), confirming independence from capture-view positions.

\subsection{Zero-inference motion-heuristic baselines}
\label{app:motion-baselines}
Throughout the paper we report two zero-inference ($\mathrm{model\_inferences}=0$) motion heuristics under identical PF41 step-budget accounting (invalid moves preserve pose and consume 1 step).

\paragraph{Local random walk.}
This 6-DoF $\mathrm{SE}(3)$ heuristic samples a camera-local isotropic 3D unit-sphere translation direction with step length $r\sim\mathrm{Uniform}(0, D_{\mathrm{scene}})$ ($D_{\mathrm{scene}}$ is the scene AABB diagonal) and an isotropic 3D rotation axis $\bm{\omega}\sim\mathcal{U}(S^2)$ with angle $\Delta\theta\sim\mathrm{Uniform}(-180^\circ,+180^\circ)$.

\paragraph{Yaw scan.}
This zero-translation heuristic sweeps horizontal yaw within the $\pm 15^\circ$ local field at $5^\circ$/step around $+Y$, isolating the aesthetic gain achievable by in-place camera panning without 3D spatial repositioning.

Table~\ref{tab:app-a3-motion} reports $G_{\rm best}[s_1]$, $\operatorname{AUC}[s_1]$, and $\operatorname{SR}_{0.3}[s_1]$ under the standard full-benchmark protocol ($T=40$; $2\text{k}$ checkpoint).

\begin{table*}[t]
\centering\renewcommand{\arraystretch}{1.08}\small
\setlength{\tabcolsep}{5.5pt}
\caption{\textbf{Zero-inference motion heuristics and execution strides under the standard full-benchmark protocol ($T=40$; $2\text{k}$ checkpoint).}}
\label{tab:app-a3-motion}
\resizebox{0.84\textwidth}{!}{%
\begin{tabular}{@{}lrrrrrr@{}}\toprule
& \multicolumn{3}{c}{All Worlds 49 ($n=294$)} & \multicolumn{3}{c}{Unseen 36 ($n=216$)} \\
\cmidrule(lr){2-4}\cmidrule(l){5-7}
Model & $G_{\rm best}[s_1]$ & $\operatorname{AUC}[s_1]$ & $\operatorname{SR}_{0.3}$(\%)
      & $G_{\rm best}[s_1]$ & $\operatorname{AUC}[s_1]$ & $\operatorname{SR}_{0.3}$(\%) \\\midrule
EZCAM-Policy ($m=1$, default) & 0.5126 & 0.3857 & 57.8 & 0.3985 & 0.3006 & 47.2 \\
EZCAM-Policy ($m=4$, frozen) & \textbf{0.5407} & \textbf{0.3886} & 57.1 & 0.4249 & 0.3105 & 49.1 \\
EZCAM-Policy ($m=8$, frozen) & 0.5336 & 0.3848 & \textbf{58.8} & \textbf{0.4287} & \textbf{0.3159} & \textbf{50.5} \\
DualVLN & 0.2817 & 0.2438 & 30.6 & 0.1094 & 0.1040 & 13.9 \\
StreamVLN & 0.2637 & 0.2159 & 32.0 & 0.2054 & 0.1776 & 23.6 \\\midrule
Local random walk & 0.1199 & 0.0735 & 13.6 & 0.0898 & 0.0498 & 11.1 \\
Yaw scan & 0.2365 & 0.2189 & 31.0 & 0.2068 & 0.1911 & 26.4 \\\bottomrule
\end{tabular}%
}
\end{table*}

\paragraph{Zero-shot VLM photographer and online score-guided search baselines.}
To test whether closed-loop active photography can be solved by zero-shot vision-language reasoning or online score-guided local search without policy training, Table~\ref{tab:app-b2-vlm-search} (Appendix~\ref{app:additional-tables}) evaluates a closed-loop \textbf{Qwen3-VL-8B Photographer} (proposing bounded 6-DoF camera moves directly from ego-centric RGB observations) and online $s_1$-guided \textbf{Hill-Climbing / $(1+1)$-ES} (querying $s_1$ after each executed perturbation and charging reverse retreat moves under the identical 40-request no-preview budget) under the 30-scene unseen diagnostic setting of Table~\ref{tab:policy-unseen30-k100} ($n=180$, $T=40$, visible-subject starts with mean $s_1(I_0)=4.84$ and collision-to-rotation recovery).
Both zero-shot VLM reasoning ($G_{\rm best}[s_{\mathrm{indep}}]=0.1839$) and online $s_1$ hill-climbing ($0.1438$) lag far behind \textbf{EZCAM-Policy (6k)} ($0.4118$), empirically confirming Insight~1: local greedy search expends budget retreating from non-monotonic transit drops and stalls in local valleys, whereas 32-step chunk prediction crosses them smoothly.

\subsection{Held-out scorer replay and Top-\texorpdfstring{$K$}{K} onboard shutter buffer}
\label{app:heldout-shutter}
Primary closed-loop benchmark tables (Tables~\ref{tab:policy-main}, \ref{tab:policy-data}, and~\ref{tab:policy-horizon}) report fast two-head LAION V1 scores $s_1(I)$ (averaging OpenCLIP ViT-L/14 and ViT-B/32 aesthetic heads), matching the frozen internal score feature of EZCAM-Policy.
To verify generalization across unseen scorer architectures under the standard full-benchmark protocol ($2\text{k}$ checkpoint, uncurated geometric starts with Unseen mean $s_1(I_0)=4.38$, and strict hard collision), we deterministically replay all $194{,}070$ active unique $(e, \text{SHA-256})$ poses ($206{,}273$ cached frame evaluations across three shard groups) with zero pixel-hash mismatches ($\text{hash failures}=0$) and evaluate them with held-out SigLIP-SO400M V2.5 ($s_{2.5}(I)=\operatorname{clip}(s_{2.5}^{\mathrm{raw}}(I),0,10)$), held-out MLLM ArtiMuse ($s_{\mathrm{AM}}(I)=s_{\mathrm{AM}}^{\mathrm{raw}}(I)/10$), the three-model ensemble $s_{\mathrm{ens}}(I)=\frac{1}{3}[s_1(I)+s_{2.5}(I)+s_{\mathrm{AM}}(I)]$ (Eq.~\ref{eq:scorer-normalized-mean}), and the strictly LAION-free ensemble $s_{\mathrm{indep}}(I)=\frac{1}{2}[s_{2.5}(I)+s_{\mathrm{AM}}(I)]$.

Tables~\ref{tab:app-a1-all49}--\ref{tab:app-a1-unseen30-strict} report the complete held-out cross-scorer results across All Worlds 49 ($n=294$), Unseen 36 ($n=216$), and same-protocol Unseen 30 ($n=180$, excluding six compact object shells): \textbf{EZCAM-Policy} ranks first across every held-out scorer ($s_{2.5}$, $s_{\mathrm{AM}}$, $s_{\mathrm{indep}}$), achieving $G_{\rm best}[s_{\mathrm{indep}}]$ of \textbf{0.5213} (All Worlds 49), \textbf{0.3758} (Unseen 36), and \textbf{0.4241} (same-protocol Unseen 30), with $\operatorname{AUC}[s_{\mathrm{indep}}]$ of \textbf{0.3746}, \textbf{0.2697}, and \textbf{0.3034}.
By contrast, terminal non-shutter gains $G_{\rm end}[s_{\mathrm{indep}}]$ of Local random walk collapse to $-0.5541$ (All Worlds 49) and $-0.7292$ (Unseen 36) because undirected 6-DoF diffusion drifts away after passing incidental views.

\paragraph{Top-$K$ onboard distinct-view buffer.}
For Table~\ref{tab:policy-heldout-shutter} in the main text, the onboard camera maintains a compact burst buffer $\mathcal{B}_K$ of up to $K\in\{1,3,5,10\}$ distinct views (deduplicated by exact \texttt{rgb\_sha256}) ranked solely by onboard $s_1(I_t)$ over $t\in\{0,\dots,40\}$, evaluated under held-out $s_{\mathrm{indep}}$ via $G_{\rm sel}[s_{\mathrm{indep}}]=\max_{t\in\mathcal{B}_K} s_{\mathrm{indep}}(I_t)-s_{\mathrm{indep}}(I_0)$, where $K=1$ corresponds to single-shot $\arg\max_t s_1(I_t)$.
Evaluated under identical PF41 starts and buffer accounting against StreamVLN, DualVLN, and Yaw scan across All Worlds 49 ($n=294$), Unseen 36 ($n=216$), and same-protocol Unseen 30 ($n=180$, Unseen 36 excluding six compact object shells; EZCAM 41-frame $s_{\mathrm{indep}}$ peak $0.4241$, $\operatorname{SR}_{0.3}=45.6\%$), \textbf{EZCAM-Policy} outperforms every baseline at every quota $K\in\{1,3,5,10\}$ on all three splits.
Even at single-shot $K=1$, \textbf{EZCAM-Policy} achieves $2.0\text{--}4.8\times$ the independent gain of StreamVLN and Yaw scan across splits (while DualVLN degrades below zero on unseen environments); retaining a compact 3-to-5 shot burst ($K=3,5$) already surpasses the full 41-frame peak of every competing baseline, reaching $83.6\%\text{--}87.3\%$ of EZCAM-Policy's own 41-frame ceiling at $K=10$ and raising $\operatorname{SR}_{0.3}$ to $44.6\%$ on All Worlds 49 ($51.0\%$ peak) and $41.1\%$ on Unseen 30 ($45.6\%$ peak).

\begin{table}[!htb]
\centering\renewcommand{\arraystretch}{1.08}\small
\setlength{\tabcolsep}{5.5pt}
\caption{\textbf{Held-out cross-scorer breakdown on All Worlds 49 ($n=294$, $T=40$; $2\text{k}$ checkpoint).} $s_{\mathrm{indep}}=\frac{1}{2}[s_{2.5}+s_{\mathrm{AM}}]$ excludes LAION V1 ($s_1$).}
\label{tab:app-a1-all49}
\resizebox{\linewidth}{!}{%
\begin{tabular}{@{}lrrrrrr@{}}\toprule
Model & $G_{\rm best}[s_1]$ & $G_{\rm best}[s_{2.5}]$ & $G_{\rm best}[s_{\mathrm{AM}}]$ & $G_{\rm best}[s_{\mathrm{indep}}]$ & $\operatorname{AUC}[s_{\mathrm{indep}}]$ & $\operatorname{SR}_{0.3}[s_{\mathrm{indep}}]$(\%) \\\midrule
\textbf{EZCAM-Policy (ours)} & \textbf{0.5126} & \textbf{0.3648} & \textbf{0.8854} & \textbf{0.5213} & \textbf{0.3746} & \textbf{51.0} \\
DualVLN & 0.2817 & 0.2060 & 0.4369 & 0.2599 & 0.2222 & 23.1 \\
StreamVLN & 0.2637 & 0.2359 & 0.6574 & 0.3550 & 0.2644 & 32.0 \\
Uni-NaVid & 0.2302 & 0.2140 & 0.4830 & 0.2720 & 0.2139 & 29.9 \\
NavGPT-2 & 0.2171 & 0.2190 & 0.4768 & 0.2671 & 0.2104 & 24.1 \\
NaviLLM & 0.2077 & 0.1865 & 0.5317 & 0.2918 & 0.2257 & 26.9 \\
NaVid & 0.2006 & 0.1992 & 0.4887 & 0.2701 & 0.2040 & 27.2 \\\midrule
Local random walk & 0.1199 & 0.1477 & 0.3111 & 0.1899 & 0.1105 & 18.4 \\
Yaw scan & 0.2365 & 0.1841 & 0.4635 & 0.2658 & 0.2457 & 31.3 \\\bottomrule
\end{tabular}%
}
\end{table}

\begin{table}[!htb]
\centering\renewcommand{\arraystretch}{1.08}\small
\setlength{\tabcolsep}{5.5pt}
\caption{\textbf{Held-out cross-scorer breakdown on Unseen 36 scenes ($n=216$, $T=40$; $2\text{k}$ checkpoint).}}
\label{tab:app-a1-unseen36}
\resizebox{\linewidth}{!}{%
\begin{tabular}{@{}lrrrrrrr@{}}\toprule
Model & $G_{\rm best}[s_1]$ & $\operatorname{SR}_{0.3}[s_1]$(\%) & $G_{\rm best}[s_{2.5}]$ & $G_{\rm best}[s_{\mathrm{AM}}]$ & $G_{\rm best}[s_{\mathrm{indep}}]$ & $\operatorname{AUC}[s_{\mathrm{indep}}]$ & $\operatorname{SR}_{0.3}[s_{\mathrm{indep}}]$(\%) \\\midrule
\textbf{EZCAM-Policy (ours)} & \textbf{0.3985} & \textbf{47.2} & \textbf{0.2784} & \textbf{0.6345} & \textbf{0.3758} & \textbf{0.2697} & \textbf{40.3} \\
DualVLN & 0.1094 & 13.9 & 0.1021 & 0.1609 & 0.0927 & 0.0874 & 8.3 \\
StreamVLN & 0.2054 & 23.6 & 0.1716 & 0.4047 & 0.2193 & 0.1668 & 19.9 \\
Uni-NaVid & 0.1602 & 21.3 & 0.1582 & 0.2835 & 0.1614 & 0.1369 & 17.1 \\
NavGPT-2 & 0.0999 & 12.0 & 0.1202 & 0.2275 & 0.1256 & 0.1011 & 12.0 \\
NaviLLM & 0.1345 & 15.7 & 0.1338 & 0.2733 & 0.1537 & 0.1294 & 13.9 \\
NaVid & 0.1386 & 13.9 & 0.1369 & 0.2600 & 0.1482 & 0.1246 & 13.4 \\\midrule
Local random walk & 0.0898 & 11.1 & 0.1291 & 0.2325 & 0.1430 & 0.0759 & 17.1 \\
Yaw scan & 0.2068 & 26.4 & 0.1673 & 0.3867 & 0.2245 & 0.2069 & 23.6 \\\bottomrule
\end{tabular}%
}
\end{table}

\begin{table}[!htb]
\centering\renewcommand{\arraystretch}{1.08}\small
\setlength{\tabcolsep}{5.5pt}
\caption{\textbf{Held-out cross-scorer breakdown on same-protocol Unseen 30 scenes} (Unseen 36 excluding six compact object shells, $n=180$, $T=40$; $2\text{k}$ checkpoint).}
\label{tab:app-a1-unseen30-strict}
\resizebox{\linewidth}{!}{%
\begin{tabular}{@{}lrrrrrrr@{}}\toprule
Model & $G_{\rm best}[s_1]$ & $\operatorname{SR}_{0.3}[s_1]$(\%) & $G_{\rm best}[s_{2.5}]$ & $G_{\rm best}[s_{\mathrm{AM}}]$ & $G_{\rm best}[s_{\mathrm{indep}}]$ & $\operatorname{AUC}[s_{\mathrm{indep}}]$ & $\operatorname{SR}_{0.3}[s_{\mathrm{indep}}]$(\%) \\\midrule
\textbf{EZCAM-Policy (ours)} & \textbf{0.4322} & \textbf{51.1} & \textbf{0.3059} & \textbf{0.7223} & \textbf{0.4241} & \textbf{0.3034} & \textbf{45.6} \\
DualVLN & 0.1198 & 15.6 & 0.1121 & 0.1837 & 0.1034 & 0.0972 & 8.9 \\
StreamVLN & 0.2181 & 25.0 & 0.1906 & 0.4653 & 0.2516 & 0.1890 & 23.3 \\
Uni-NaVid & 0.1658 & 23.3 & 0.1778 & 0.3148 & 0.1788 & 0.1498 & 18.3 \\
NavGPT-2 & 0.1107 & 13.3 & 0.1381 & 0.2638 & 0.1443 & 0.1152 & 13.9 \\
NaviLLM & 0.1374 & 15.6 & 0.1429 & 0.3080 & 0.1700 & 0.1413 & 15.0 \\
NaVid & 0.1460 & 13.9 & 0.1483 & 0.2966 & 0.1681 & 0.1401 & 15.6 \\\midrule
Local random walk & 0.0717 & 8.9 & 0.1211 & 0.2454 & 0.1468 & 0.0795 & 17.8 \\
Yaw scan & 0.1963 & 23.3 & 0.1711 & 0.4265 & 0.2452 & 0.2262 & 27.8 \\\bottomrule
\end{tabular}%
}
\end{table}

\subsection{External third-party IQA evaluation}
\label{app:external-iqa}
To test whether closed-loop gains transfer beyond our scorer suite, we replay All 49 episodes ($n=294$) under four independent off-the-shelf image quality/aesthetics evaluators in their native scales: Q-Align~\citep{wu2024qalign} Quality and Aesthetics, MUSIQ~\citep{ke2021musiq}, and NIMA~\citep{talebi2018nima}.
Table~\ref{tab:app-b3-iqa} reports mean gains $\overline{\Delta}$ from $I_0$ to the best-observed frame across all nine methods.
\textbf{EZCAM-Policy} leads every baseline on all four evaluators.

\begin{table}[!htb]
\centering\renewcommand{\arraystretch}{1.08}\small
\setlength{\tabcolsep}{5.5pt}
\caption{\textbf{External third-party IQA evaluation on All 49 scenes ($n=294$).}
Mean gains $\overline{\Delta}$ from $I_0$ to the best-observed frame under Q-Align~\citep{wu2024qalign} (Quality / Aesthetics), MUSIQ~\citep{ke2021musiq}, and NIMA~\citep{talebi2018nima} in native scales.}
\label{tab:app-b3-iqa}
\resizebox{\linewidth}{!}{%
\begin{tabular}{@{}lrrrr@{}}\toprule
Model & $\overline{\Delta}$ Q-Align$_{\mathrm{Q}}$ & $\overline{\Delta}$ Q-Align$_{\mathrm{A}}$ & $\overline{\Delta}$ MUSIQ & $\overline{\Delta}$ NIMA \\\midrule
\textbf{EZCAM-Policy (ours)} & \textbf{+0.1628} & \textbf{+0.1508} & \textbf{+3.0445} & \textbf{+0.1244} \\
DualVLN & $+0.0637$ & $+0.0964$ & $+1.1182$ & $+0.1035$ \\
StreamVLN & $+0.0439$ & $+0.0367$ & $+0.7249$ & $+0.0080$ \\
Uni-NaVid & $+0.0264$ & $+0.0194$ & $+0.5151$ & $+0.0185$ \\
NavGPT-2 & $+0.0543$ & $+0.0579$ & $+1.0180$ & $+0.0848$ \\
NaviLLM & $+0.0346$ & $+0.0205$ & $+0.7296$ & $+0.0279$ \\
NaVid & $+0.0046$ & $-0.0008$ & $+0.2086$ & $+0.0151$ \\\midrule
Local random walk & $+0.0520$ & $+0.0421$ & $+0.9196$ & $+0.0412$ \\
Yaw scan & $+0.0145$ & $+0.0194$ & $+0.2966$ & $+0.0158$ \\\bottomrule
\end{tabular}%
}
\end{table}

\subsection{Controlled supervision and policy ablations}
\label{app:policy-ablations}
\paragraph{Matched continuation.}
The reference uniform and branch continuations share the base checkpoint,
backbone, normalizer, optimizer schedule, batch size, and update count (full architecture, two-stage optimization schedule, and explicit branch-selection objective in Appendix~\ref{app:additional-tables}, Table~\ref{tab:app-hyperparams} and Eq.~\ref{eq:branch-continuation}).
Uniform uses recorded trajectory windows; branch supervision selects targets from the highest-return sampled continuation within the $k=32$ horizon (Eq.~\ref{eq:branch-continuation}). The encoder receives the observed window, and branch targets supervise the predicted action chunk, providing cleaner multi-step supervision across local score fluctuations.

Both continuation settings use the same 294 starts, initial scores, renderer,
scorer, action limits, and 40-request/41-observation budget.

\paragraph{Results on all 49 scenes.}
Under this shared budget ($T=40$), full EZCAM-Policy with branch-selected continuation achieves $G_{\rm best}=0.5126$, $\operatorname{AUC}=0.3857$, and $\operatorname{SR}_{0.3}=57.8\%$, whereas uniform window continuation achieves $G_{\rm best}=0.4022$ and $\operatorname{AUC}=0.3208$ (Table~\ref{tab:policy-ablation-49}).

\begin{table}[!htb]
\centering\renewcommand{\arraystretch}{1.08}\small
\caption{\textbf{Ablation on branch-selected continuation across all 49 scenes ($n=294$, $T=40$).}}
\label{tab:policy-ablation-49}
\resizebox{\linewidth}{!}{%
\begin{tabular}{@{}llrr@{}}\toprule
Supervision & Setting & $G_{\rm best}\uparrow$ & $\operatorname{AUC}\uparrow$ \\\midrule
Branch-selected continuation & Full EZCAM-Policy & \textbf{0.5126} & \textbf{0.3857} \\
Uniform window continuation & Same budget & 0.4022 & 0.3208 \\\bottomrule
\end{tabular}%
}
\end{table}

\begin{figure*}[t]
\centering
\includegraphics[width=0.94\textwidth]{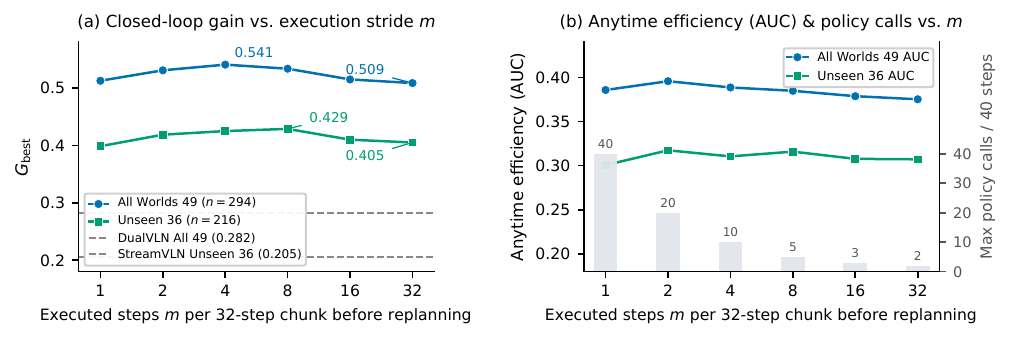}
\caption{\textbf{Test-time execution stride ablation on the frozen $2\text{k}$ EZCAM-Policy checkpoint ($k=32$, $T=40$).} \emph{(a)} $G_{\rm best}$ across strides $m\in\{1,2,4,8,16,32\}$ on All Worlds 49 ($n=294$) and Unseen 36 ($n=216$), with external VLN baselines shown as dashed references. \emph{(b)} Anytime efficiency ($\operatorname{AUC}$, curves) and maximum policy forward calls per 40-step episode (bars) vs.\ $m$.}
\label{fig:app-b4-stride}
\end{figure*}

\paragraph{Disentangling 32-step supervision from replanning frequency ($m\in\{1,\dots,32\}$).}
To test whether the gains of $k=32$ stem from noise-filtered multi-step trajectory supervision or merely from $m=1$ step-by-step reactive replanning, we freeze the EZCAM-Policy checkpoint from Table~\ref{tab:policy-main} ($k=32$, $T=40$) and vary only the test-time execution stride $m\in\{1,2,4,8,16,32\}$ (Figure~\ref{fig:app-b4-stride} and Table~\ref{tab:app-b4-stride}).
Each forward pass predicts 32 future 10-D camera increments and executes up to $m$ increments sequentially (appending each observed view and executed action into the history window and charging one step of the $T=40$ budget), flushing any remaining queued actions to replan early only when an action is rejected by collision checking.
Setting $m=1$ reproduces Table~\ref{tab:policy-main} with zero numerical discrepancy ($G_{\rm best}=0.5126$, $\operatorname{AUC}=0.3857$, $\operatorname{SR}_{0.3}=57.8\%$ on All 49; $0.3985$ on Unseen 36).

Two findings strengthen Insight 1:
\emph{(i) Coherent 32-step trajectory prior:} Even at $m=32$ (executing the entire 32-step predicted chunk with at most 2 forward passes over $T=40$ steps), EZCAM-Policy achieves $G_{\rm best}=0.5086$ on All Worlds 49 and $0.4050$ on Unseen 36 (surpassing $m=1$'s $0.3985$ on Unseen 36 and far exceeding $k=1$ short-horizon BC and all VLN baselines). This confirms that 32-step chunk supervision eliminates single-step aesthetic score jitter and learns coherent 3D framing paths.
\emph{(ii) Multi-step execution sweet spot and $4\text{--}8\times$ speedup:} Moderate strides ($m=4$ on All Worlds 49: $0.5407$; $m=8$ on Unseen 36: $0.4287$, $\operatorname{SR}_{0.3}=50.5\%$) outperform $m=1$ while reducing policy forward calls by $4\text{--}8\times$ by suppressing step-to-step stochastic resampling jitter.

\begin{table*}[t]
\centering\renewcommand{\arraystretch}{1.08}\small
\setlength{\tabcolsep}{5.5pt}
\caption{\textbf{Test-time execution stride $m\in\{1,2,4,8,16,32\}$ on the frozen Table~\ref{tab:policy-main} $2\text{k}$ checkpoint ($k=32$, PF41 budget $T=40$).} $\Delta G_{\rm best}$ is relative to the paper default $m=1$.}
\label{tab:app-b4-stride}
\resizebox{0.92\textwidth}{!}{%
\begin{tabular}{@{}lrrrrrrrrr@{}}\toprule
& & \multicolumn{4}{c}{All Worlds 49 ($n=294$)} & \multicolumn{4}{c}{Unseen 36 ($n=216$)} \\
\cmidrule(lr){3-6}\cmidrule(l){7-10}
Execution Stride $m$ & Max Calls & $G_{\rm best}\uparrow$ & $\operatorname{AUC}\uparrow$ & $\operatorname{SR}_{0.3}$(\%)$\uparrow$ & $\Delta G_{\rm best}$
                     & $G_{\rm best}\uparrow$ & $\operatorname{AUC}\uparrow$ & $\operatorname{SR}_{0.3}$(\%)$\uparrow$ & $\Delta G_{\rm best}$ \\\midrule
$m=1$ (paper default) & 40 & 0.5126 & 0.3857 & 57.8 & $0.0000$ & 0.3985 & 0.3006 & 47.2 & $0.0000$ \\
$m=2$                 & 20 & 0.5308 & \textbf{0.3957} & \textbf{59.5} & $+0.0183$ & 0.4186 & \textbf{0.3174} & 50.0 & $+0.0201$ \\
$m=4$                 & 10 & \textbf{0.5407} & 0.3886 & 57.1 & \textbf{+0.0281} & 0.4249 & 0.3105 & 49.1 & $+0.0264$ \\
$m=8$                 & 5  & 0.5336 & 0.3848 & 58.8 & $+0.0211$ & \textbf{0.4287} & 0.3159 & \textbf{50.5} & \textbf{+0.0302} \\
$m=16$                & 3  & 0.5149 & 0.3787 & 58.8 & $+0.0023$ & 0.4099 & 0.3077 & 50.0 & $+0.0114$ \\
$m=32$ (full chunk)   & 2  & 0.5086 & 0.3752 & 57.8 & $-0.0040$ & 0.4050 & 0.3072 & 49.1 & $+0.0065$ \\\bottomrule
\end{tabular}%
}
\end{table*}

\paragraph{Controlled horizon isolation ($k\in\{1,\dots,32\}$) and scaling ceiling.}
In Table~\ref{tab:policy-horizon} of the main text, all six horizons $k\in\{1,2,4,8,16,32\}$ share a matched $2\text{k}$-step schedule on a unified dense 32-step window cache to strictly isolate prediction horizon length, demonstrating both a $+0.1162$ advantage of $k=32$ over $k=1$ ($0.5711$ vs.\ $0.4549$, paired 95\% bootstrap CI $[-0.1775,-0.0579]$) and a higher $0.5711 / 65.0\%$ performance ceiling under denser windowing alongside our standard two-stage multi-source checkpoint in Tables~\ref{tab:policy-main} and~\ref{tab:policy-data} ($0.5126 / 0.3857 / 57.8\%$).

\paragraph{Detailed tabular results on the 30-scene long-horizon diagnostic ($T=1\dots 100$).}
Tables~\ref{tab:policy-unseen30-k100} and~\ref{tab:policy-ablations} provide the complete numerical readouts corresponding to Figure~\ref{fig:policy-unseen30-external}a,b of the main text across 30 unseen room and outdoor scenes ($n=180$, visible-subject starts with mean $s_1(I_0)=4.84$).
To prevent deterministic navigation baselines from stalling indefinitely against walls over $T=100$ steps and grant competing methods the most favorable long-horizon operating regime, this diagnostic enables symmetric collision-to-rotation recovery (retaining valid rotational increments when translation is blocked); even under this baseline-favorable recovery setting, both \textbf{EZCAM-Policy ($2\text{k}$)} and full-epoch \textbf{($6\text{k}$)} lead every baseline across all step budgets $T\in\{10,20,40,100\}$.

\begin{table}[!htb]
\centering\renewcommand{\arraystretch}{1.08}\small
\setlength{\tabcolsep}{5.5pt}
\caption{\textbf{Numerical results for the long-horizon 30-scene diagnostic} ($n=180$; companion to Figure~\ref{fig:policy-unseen30-external}b), comparing both \textbf{EZCAM-Policy (2k)} and full-epoch \textbf{(6k)} checkpoints against external baselines.}
\label{tab:policy-unseen30-k100}
\resizebox{\linewidth}{!}{%
\begin{tabular}{@{}llrrrrrrr@{}}\toprule
& & \multicolumn{4}{c}{Best-Observed Gain $G_{\rm best}@T\uparrow$} & \multicolumn{3}{c}{Anytime Efficiency $\operatorname{AUC}@T\uparrow$}\\
\cmidrule(lr){3-6}\cmidrule(l){7-9}
Category & Method & $T=10$ & $T=20$ & $T=40$ & $T=100$ & $\operatorname{AUC}@10$ & $\operatorname{AUC}@40$ & $\operatorname{AUC}@100$\\\midrule
\textbf{Embodied Photography} & \textbf{EZCAM-Policy (2k)} & \textbf{0.2621} & 0.3123 & 0.3686 & 0.4266 & \textbf{0.1842} & 0.2912 & 0.3584\\
& \textbf{EZCAM-Policy (6k)} & 0.2360 & \textbf{0.3192} & \textbf{0.4081} & \textbf{0.5068} & 0.1649 & \textbf{0.2994} & \textbf{0.3956}\\\midrule
Trajectory Generation & GenDoP~\citep{zhang2025gendop} & 0.0736 & 0.1204 & 0.1926 & 0.2696 & 0.0536 & 0.1193 & 0.1943\\
& GenDoP (replan every 10 steps) & 0.0736 & 0.1076 & 0.1413 & 0.2163 & 0.0536 & 0.1006 & 0.1524\\\midrule
Embodied VLN Adapters & StreamVLN~\citep{wei2025streamvln} & 0.1923 & 0.2309 & 0.2725 & 0.3148 & 0.1486 & 0.2194 & 0.2657\\
& NaviLLM~\citep{zheng2024navillm} & 0.1717 & 0.2079 & 0.2405 & 0.2843 & 0.1335 & 0.1954 & 0.2371\\
& DualVLN~\citep{wei2025dualvln} & 0.1478 & 0.1839 & 0.2273 & 0.2784 & 0.1147 & 0.1765 & 0.2246\\
& Uni-NaVid~\citep{zhang2025uninavid} & 0.1669 & 0.1943 & 0.2275 & 0.2655 & 0.1288 & 0.1843 & 0.2281\\
& NaVid~\citep{zhang2024navid} & 0.1600 & 0.1953 & 0.2240 & 0.2464 & 0.1252 & 0.1841 & 0.2151\\
& NavGPT-2~\citep{zhou2024navgpt2} & 0.1585 & 0.1916 & 0.2104 & 0.2407 & 0.1204 & 0.1772 & 0.2095\\\midrule
Motion Heuristics & Local random walk & 0.0533 & 0.0653 & 0.1011 & 0.1412 & 0.0296 & 0.0623 & 0.0988 \\
& Yaw scan & 0.1968 & 0.1968 & 0.1968 & 0.1968 & 0.1467 & 0.1843 & 0.1918\\\bottomrule
\end{tabular}%
}
\end{table}

\begin{table}[!htb]
\centering\renewcommand{\arraystretch}{1.08}\small
\setlength{\tabcolsep}{5.5pt}
\caption{\textbf{Numerical results for EZCAM-Policy ablations on 30 unseen scenes ($n=180$; companion to Figure~\ref{fig:policy-unseen30-ablations}a).}
The four ablated rows share a matched $2\text{k}$-step optimization budget with \textbf{EZCAM-Policy (2k)}, shown alongside full-epoch \textbf{EZCAM-Policy (6k)}.}
\label{tab:policy-ablations}
\resizebox{\linewidth}{!}{%
\begin{tabular}{@{}lrrrrrrr@{}}\toprule
& \multicolumn{4}{c}{Best-Observed Gain $G_{\rm best}@T\uparrow$} & \multicolumn{3}{c}{Anytime Efficiency $\operatorname{AUC}@T\uparrow$}\\
\cmidrule(lr){2-5}\cmidrule(l){6-8}
Variant & $T=10$ & $T=20$ & $T=40$ & $T=100$ & $\operatorname{AUC}@10$ & $\operatorname{AUC}@40$ & $\operatorname{AUC}@100$\\\midrule
\textbf{EZCAM-Policy (2k)} & \textbf{0.2621} & 0.3123 & 0.3686 & 0.4266 & \textbf{0.1842} & 0.2912 & 0.3584\\
\textbf{EZCAM-Policy (6k)} & 0.2360 & \textbf{0.3192} & \textbf{0.4081} & \textbf{0.5068} & 0.1649 & \textbf{0.2994} & \textbf{0.3956}\\\midrule
w/o aesthetic pretraining (2k) & 0.2355 & 0.2667 & 0.3044 & 0.4028 & 0.1666 & 0.2495 & 0.3124\\
w/o $s_1$ score features (2k) & 0.2042 & 0.2500 & 0.2770 & 0.3475 & 0.1433 & 0.2294 & 0.2816\\
frozen score projection (2k) & 0.1906 & 0.2400 & 0.2812 & 0.3319 & 0.1376 & 0.2241 & 0.2755\\
Internet only (2k) & 0.1889 & 0.2489 & 0.2959 & 0.3798 & 0.1358 & 0.2282 & 0.2984\\\bottomrule
\end{tabular}%
}
\end{table}

\subsection{Supplementary feature-adapter records}
\label{app:adapter-results}
The six VLN adapters in Table~\ref{tab:policy-main} pair frozen public visual backbones from
DualVLN~\citep{wei2025dualvln}, StreamVLN~\citep{wei2025streamvln},
Uni-NaVid~\citep{zhang2025uninavid}, NavGPT-2~\citep{zhou2024navgpt2},
NaviLLM~\citep{zheng2024navillm}, and NaVid~\citep{zhang2024navid}
with a shared EZCAM 10-D MLP action head trained on the 75,486 Game/3DGS demonstration split (30,600 3DGS + 44,886 game) under identical PF41 starts, renderer, evaluator, and step budgets.
Table~\ref{tab:app-c2-matched-budget} (Appendix~\ref{app:additional-tables}) further evaluates the same six frozen-backbone 10-D adapters trained under the \textbf{matched multi-source schedule} of EZCAM-Policy ($519$ SpatialVID Internet updates + $2{,}000$ Game/3DGS updates, global batch size $512$).
Because VLN adapters freeze their navigation visual backbones and optimize only a shallow 10-D MLP head, pretraining that head on relative-scale Internet clips mildly dilutes metric 6-DoF step regression relative to pure Game/3DGS training in Table~\ref{tab:policy-main} (e.g., DualVLN $0.2349$ vs.\ $0.2817$; StreamVLN $0.2602$ vs.\ $0.2637$), confirming that Table~\ref{tab:policy-main} already evaluates the six VLN baselines under their stronger pure-metric configuration.
Under identical multi-source data and optimization budget, \textbf{EZCAM-Policy} ($0.5126$ $G_{\rm best}$, $0.3857$ $\operatorname{AUC}$, $\operatorname{SR}_{0.3}=57.8\%$, and $\operatorname{SR}_{0.5}=37.4\%$) outperforms the strongest VLN adapter (StreamVLN: $0.2602 / 0.2102 / 31.0\% / 19.4\%$) by nearly $2\times$ across every metric.

\subsection{Live game-engine and real-world drone deployment}
\label{app:real-deployment}
\begin{figure*}[t]
\centering
\includegraphics[width=\textwidth]{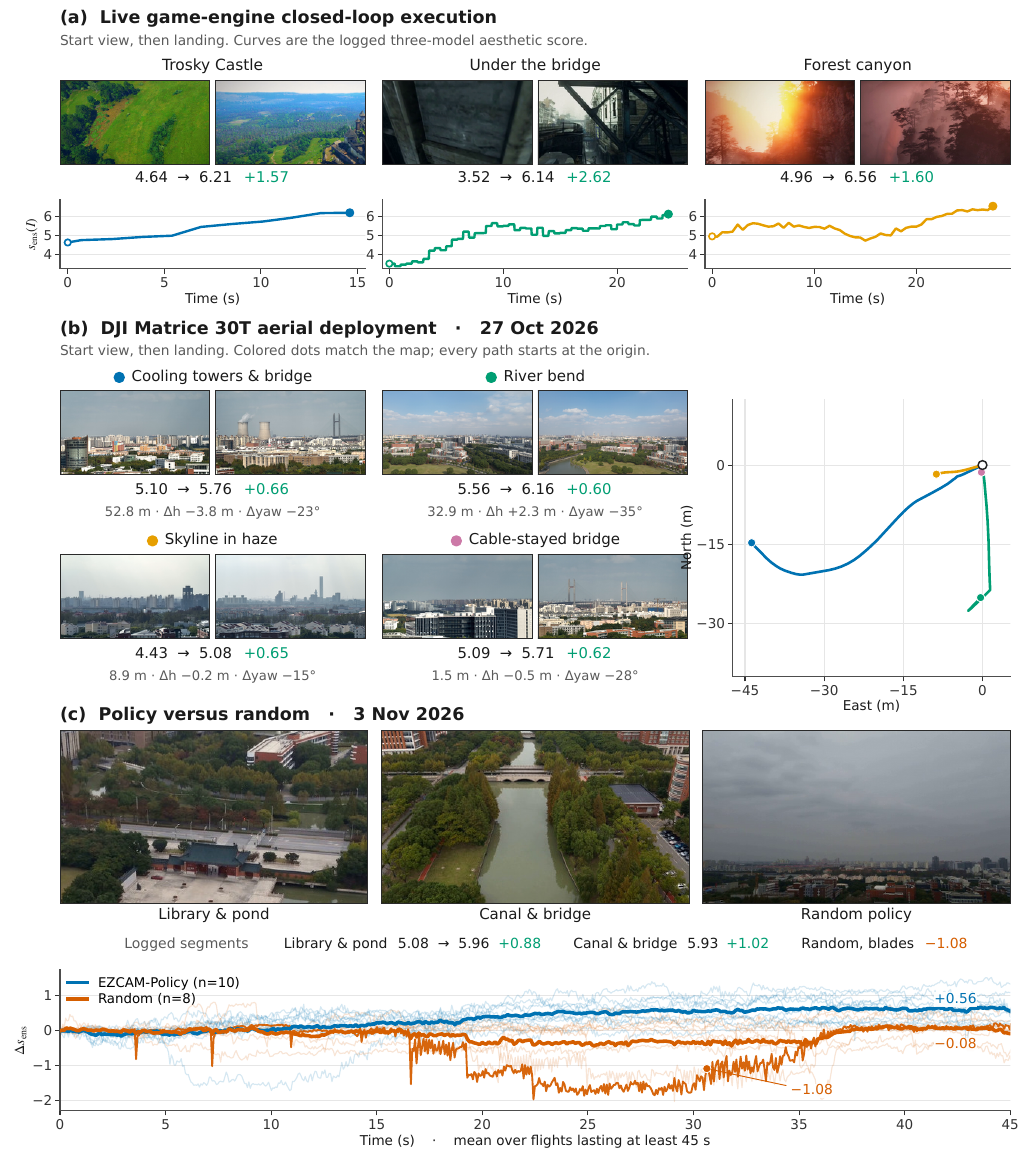}
\caption{\textbf{Zero-shot deployment gallery in live game engines~(a), multi-scale DJI Matrice 30T flights~(b), and controlled policy-vs.-random campus flights~(c; unsmoothed continuous $\Delta s_{\mathrm{ens}}(t)$ traces over $n=10$ policy and $n=8$ random flights reaching $45$\,s).}}
\label{fig:real-deployment}
\end{figure*}

Figure~\ref{fig:real-deployment} and Tables~\ref{tab:suppl-live-games}--\ref{tab:suppl-drone-flights} document the live game-engine and physical DJI Matrice 30T aerial drone deployments summarized in Section~\ref{sec:real-deployment} and Table~\ref{tab:real-world-deployment}.

\paragraph{Live game-engine closed-loop execution.}
We deploy \textbf{EZCAM-Policy} directly inside three live game engines (\emph{Kingdom Come: Deliverance II} [KCD2, 9 episodes], \emph{Black Myth: Wukong} [2 episodes], and \emph{RE9} [5 episodes]) without pre-scanned 3DGS geometry or candidate-view previews (Table~\ref{tab:suppl-live-games}).
In KCD2, each episode executes closed-loop camera actions from its initialized spawn viewpoint (with episode \texttt{TROS\_R04\_T01} recorded from step~4 after initial camera spawn settle), traversing a mean 3D path length of $6.90$\,m (net displacement $6.70$\,m, elevation change $-5.00$\,m to $+6.28$\,m, pitch change $+24.0^\circ$ to $+48.0^\circ$) and achieving a $+1.30$ mean $s_{\mathrm{ens}}$ gain ($4.45 \to 5.75$).
Across all 16 live-engine episodes, \textbf{EZCAM-Policy} achieves positive $s_{\mathrm{ens}}$ gains on \textbf{100\%} of runs, with an overall mean $\Delta s_{\mathrm{ens}}$ of \textbf{$+1.35$} (median $+1.23$, range $+0.80$ to $+2.62$).

\begin{table}[!htb]
\centering\renewcommand{\arraystretch}{1.08}\small
\setlength{\tabcolsep}{5.5pt}
\caption{\textbf{Per-episode zero-shot closed-loop execution in live game engines ($n=16$).}}
\label{tab:suppl-live-games}
\resizebox{\linewidth}{!}{%
\begin{tabular}{@{}lllrrr@{}}\toprule
Engine & Scene / Episode & Span & Start $s_{\mathrm{ens}}(I_0)$ & Peak / End $s_{\mathrm{ens}}(I)$ & Gain $\Delta s_{\mathrm{ens}}\uparrow$\\\midrule
KCD2 & Trosky Castle (\texttt{TROS\_R01\_T03}) & 10 steps (14.6\,s) & 4.64 & 6.21 & +1.57\\
KCD2 & Village terraces (\texttt{KCD2\_R01\_T01}) & 12 steps (19.7\,s) & 3.97 & 5.80 & +1.83\\
KCD2 & Town in amber light (\texttt{KCD2\_R02\_T05}) & 11 steps (15.8\,s) & 4.59 & 5.55 & +0.96\\
KCD2 & Golden ruin (\texttt{KCD2\_R08\_T03}) & 8 steps (11.4\,s) & 4.36 & 5.64 & +1.29\\
KCD2 & Gorge and fortress (\texttt{TROS\_R04\_T01}) & 6 steps (9.6\,s) & 4.94 & 6.03 & +1.10\\
KCD2 & River valley at dusk (\texttt{TROS\_R03\_T05}) & 10 steps (15.5\,s) & 3.20 & 4.81 & +1.61\\
KCD2 & Forest canopy in mist (\texttt{TROS\_R05\_T02}) & 9 steps (13.2\,s) & 4.56 & 5.64 & +1.08\\
KCD2 & Golden fields (\texttt{KCD2\_R04\_T04}) & 10 steps (14.6\,s) & 4.62 & 5.79 & +1.17\\
KCD2 & Lake and castle (\texttt{TROS\_R02\_T04}) & 10 steps (15.2\,s) & 5.14 & 6.23 & +1.09\\
Wukong & Forest canyon (\texttt{B001}) & 27.5\,s & 4.96 & 6.56 & +1.60\\
Wukong & Snow lake (\texttt{D001}) & 9.5\,s & 5.51 & 6.31 & +0.80\\
RE9 & Rain street (\texttt{N124}) & 26.0\,s & 5.98 & 7.04 & +1.06\\
RE9 & Under the bridge (\texttt{N210}) & 24.5\,s & 3.52 & 6.14 & +2.62\\
RE9 & Umbrellas (\texttt{N114}) & 15.0\,s & 5.51 & 6.60 & +1.09\\
RE9 & Brick wall (\texttt{N032}) & 14.0\,s & 4.78 & 6.14 & +1.36\\
RE9 & Steel beams (\texttt{N036}) & 6.5\,s & 4.98 & 6.31 & +1.33\\\midrule
\textbf{Mean} & \textbf{All 16 live-engine episodes} & --- & \textbf{4.70} & \textbf{6.05} & \textbf{+1.35}\\\bottomrule
\end{tabular}%
}
\end{table}

\paragraph{Physical DJI Matrice 30T aerial deployment.}
We further deploy \textbf{EZCAM-Policy} on a physical \textbf{DJI Matrice 30T} quadrotor in two outdoor flight campaigns:
\emph{(i) Multi-scale aerial photography} across five target-scene closed-loop trajectories extracted from multi-scene aerial sorties ($198.9$\,s total recorded flight stream, capturing synchronized $1920\times 1080$ wide and zoom video at $29.97$\,FPS with per-frame GPS, altitude, and gimbal telemetry scored at $15$\,Hz using a 5-sample centered median filter), achieving $+0.60$ to $+0.86$ smoothed aesthetic gain across take-off, cooling towers, urban skyline, cable-stayed bridge, and river bend scenes (mean $+0.68$; Table~\ref{tab:suppl-drone-flights}, top); and
\emph{(ii) Controlled policy-vs.-random flight evaluation} across 35 recordings ($36.3$\,min, 32,680 scored frames across four flight sessions: 16 \textbf{EZCAM-Policy} flights, 17 random-policy flights, and two stationary hovers held out), where \textbf{EZCAM-Policy} operates solely from onboard RGB observations without maps or previews and significantly outperforms the random baseline across all elapsed horizons ($15$\,s: $p=0.0066$; $30$\,s: $p=0.0013$; $45$\,s: $p=0.0025$) and first-minute best-observed gain ($+0.73$ vs.\ $+0.22$, 95\% bootstrap CI on difference $[+0.30, +0.72]$, two-sided Mann--Whitney $p=6.4\times 10^{-5}$, permutation $p=0.008$; Table~\ref{tab:suppl-drone-flights}, bottom).
Controlling for initial score $s_{\mathrm{ens}}(I_0)$ via linear covariance adjustment (ANCOVA) preserves a significant policy advantage of $+0.39$ on $G_{\rm best}@60$\,s (95\% CI $[+0.07, +0.59]$, permutation $p<10^{-4}$) and $+0.47$ on $\Delta s_{\mathrm{ens}}@30$\,s ($p=0.008$).
Sample sizes at $15$, $30$, and $45$\,s include all recordings reaching each elapsed horizon; Table~\ref{tab:suppl-drone-flights} reports 5-sample median-smoothed $\Delta s_{\mathrm{ens}}$ ($+0.62$ vs.\ $-0.05$ at $45$\,s), whereas the continuous curves in Figures~\ref{fig:policy-best-frames}\,II and~\ref{fig:real-deployment}c plot unsmoothed $\Delta s_{\mathrm{ens}}(t)$ over the identical $45$\,s cohort ($n=10$ vs.\ $n=8$; $+0.56$ vs.\ $-0.08$ at $45$\,s).

\begin{table}[!htb]
\centering\renewcommand{\arraystretch}{1.08}\small
\setlength{\tabcolsep}{5.5pt}
\caption{\textbf{Physical DJI Matrice 30T multi-scale flights (top) and controlled policy-vs.-random comparison (bottom; 5-sample median-smoothed $s_{\mathrm{ens}}(I)$).} Two-sided Mann--Whitney $p$-values compare Policy vs.\ Random.}
\label{tab:suppl-drone-flights}
\resizebox{\linewidth}{!}{%
\begin{tabular}{@{}llrrrr@{}}\toprule
Campaign / Horizon & Scene or Policy Arm & Span / Motion & Start $s_{\mathrm{ens}}$ & Peak / End $s_{\mathrm{ens}}$ & Gain $\Delta s_{\mathrm{ens}}\uparrow$\\\midrule
M30T Multi-Scale (W) & Take-off to river panorama & $11.2$\,s ($\Delta h +26.2$\,m, $\Delta\psi -28^\circ$) & 5.23 & 6.10 & +0.86\\
M30T Multi-Scale (Z) & Cooling towers \& bridge & $19.1$\,s ($52.8$\,m, $\Delta h -3.8$\,m, $\Delta\psi -23^\circ$) & 5.10 & 5.76 & +0.66\\
M30T Multi-Scale (Z) & Skyline in haze & $3.4$\,s ($8.9$\,m, $\Delta h -0.2$\,m, $\Delta\psi -15^\circ$) & 4.43 & 5.08 & +0.65\\
M30T Multi-Scale (Z) & Cable-stayed bridge & $2.3$\,s ($1.5$\,m, $\Delta h -0.5$\,m, $\Delta\psi -28^\circ$) & 5.09 & 5.71 & +0.62\\
M30T Multi-Scale (W) & River bend panorama & $12.5$\,s ($32.9$\,m, $\Delta h +2.3$\,m, $\Delta\psi -35^\circ$) & 5.56 & 6.16 & +0.60\\\midrule
Controlled Flight ($15$\,s) & \textbf{EZCAM-Policy} ($n=13$) vs.\ Random ($n=14$) & 10/13 vs.\ 6/14 pos.\ ($p=0.0066$) & --- & --- & \textbf{+0.26} vs.\ $-0.12$\\
Controlled Flight ($30$\,s) & \textbf{EZCAM-Policy} ($n=13$) vs.\ Random ($n=10$) & 10/13 vs.\ 3/10 pos.\ ($p=0.0013$) & --- & --- & \textbf{+0.46} vs.\ $-0.37$\\
Controlled Flight ($45$\,s) & \textbf{EZCAM-Policy} ($n=10$) vs.\ Random ($n=8$) & 10/10 vs.\ 5/8 pos.\ ($p=0.0025$) & --- & --- & \textbf{+0.62} vs.\ $-0.05$\\
Controlled Flight ($\le 60$\,s) & \textbf{EZCAM-Policy} ($n=16$) vs.\ Random ($n=17$) & $G_{\rm best}@60$\,s ($p=6.4\times 10^{-5}$) & 4.94 / 5.26 & 5.66 / 5.48 & \textbf{+0.73} vs.\ +0.22\\\bottomrule
\end{tabular}%
}
\end{table}

\paragraph{Deployed footprint, inference latency, and physical 6-DoF tracking accuracy.}
Table~\ref{tab:suppl-real-system} documents the complete hardware deployment specifications, runtime memory footprint, inference latency, and closed-loop physical tracking accuracy of \textbf{EZCAM-Policy} on the DJI Matrice 30T with onboard \textbf{NVIDIA Orin NX} compute:
\emph{(i) Model and memory footprint:} The deployed $2\text{k}$ checkpoint (\texttt{iter\_000002000}) contains exactly $927{,}180{,}046$ parameters (${\sim}0.927\text{B}$). Retaining the full architecture in FP16 with optimizer states stripped yields a $1.85\,\text{GB}$ ($1.73\,\text{GiB}$) checkpoint that loads in $11.3\text{--}23.4$\,s, with an offline peak allocated CUDA memory of $1.78\,\text{GiB}$ (reserved peak $1.81\,\text{GiB}$) and process resident set size (RSS) of $5.85\text{--}5.89\,\text{GiB}$. Each step takes four RGB frames resized to $224\times224$ plus four historical actions, runs 8 flow-matching sampling steps, and outputs 32 future 10-D actions, executing the first approved action before replanning.
\emph{(ii) Inference latency:} Offline inference (excluding first-step warmup) averages $0.590$\,s (median $0.592$\,s, P95 $0.610$\,s), while live tethered execution with the aircraft (including first-step warmup) averages $0.985$\,s (median $0.950$\,s, P95 $1.219$\,s).
\emph{(iii) Closed-loop dispatch and physical tracking:} From inference start, action command dispatch completes in $1.084$\,s on average (P95 $1.452$\,s) and software-confirmed physical action execution completes in $2.066$\,s (P95 $2.480$\,s). Across real-flight steps, the quadrotor and gimbal achieve a terminal 3D position error of $9.95$\,cm (P95 $27.13$\,cm) relative to the dispatched target, a camera attitude error of $0.105^\circ$ (P95 $0.228^\circ$) relative to the adapted gimbal target, and an attitude error of $0.377^\circ$ (P95 $0.827^\circ$) relative to the raw model target (including the airframe-to-gimbal attitude mapping offset).

\begin{table*}[t]
\centering\renewcommand{\arraystretch}{1.08}\small
\caption{\textbf{Hardware footprint, inference latency, and 6-DoF tracking accuracy of EZCAM-Policy on the DJI Matrice 30T.}}
\label{tab:suppl-real-system}
\resizebox{\textwidth}{!}{%
\begin{tabular}{@{}lll@{}}\toprule
Category & Metric / Specification & Verified Measurement\\\midrule
Model \& Footprint & Onboard compute module & NVIDIA Orin NX (mounted on DJI Matrice 30T)\\
& Deployed checkpoint & $2\text{k}$ checkpoint (\texttt{iter\_000002000})\\
& Parameter count & $927{,}180{,}046$ (${\sim}0.927\text{B}$ / $927\text{M}$)\\
& Inference precision & FP16 (full model retained, training optimizer states removed)\\
& Deployed checkpoint file size & $1.85\,\text{GB}$ (${\sim}1.73\,\text{GiB}$)\\
& Policy input representation & $4$ RGB images (resized to $224\times224$) $+$ $4$ historical camera actions\\
& Flow-matching sampling schedule & $8$ sampling steps per inference call\\
& Policy output \& execution rule & $32$ future actions ($10$-D each); only the first approved action is executed per step\\
& Historical model load time & $11.3\text{--}23.4$\,s\\
& Offline CUDA memory peak & ${\sim}1.78\,\text{GiB}$ allocated peak (${\sim}1.81\,\text{GiB}$ reserved cache peak)\\
& Offline process memory peak (RSS) & $5.85\text{--}5.89\,\text{GiB}$\\\midrule
Inference Latency & Offline inference (excluding first warmup) & Mean \textbf{$0.590$\,s}, median $0.592$\,s, P95 $0.610$\,s\\
& Live drone operation (including first warmup) & Mean \textbf{$0.985$\,s}, median $0.950$\,s, P95 $1.219$\,s\\\midrule
Physical Execution & Inference start $\to$ action command dispatch & Mean \textbf{$1.084$\,s}, P95 $1.452$\,s\\
\& 6-DoF Tracking & Inference start $\to$ confirmed action completion & Mean \textbf{$2.066$\,s}, P95 $2.480$\,s\\
& Terminal position error (vs.\ dispatched target) & Mean \textbf{$9.95$\,cm}, P95 $27.13$\,cm\\
& Camera attitude error (vs.\ adapted target) & Mean \textbf{$0.105^\circ$}, P95 $0.228^\circ$\\
& Camera attitude error (vs.\ raw model target, incl.\ mapping offset) & Mean \textbf{$0.377^\circ$}, P95 $0.827^\circ$\\\bottomrule
\end{tabular}%
}
\end{table*}

\subsection{Additional baseline controls and training specifications}
\label{app:additional-tables}
Tables~\ref{tab:app-b2-vlm-search}--\ref{tab:app-c2-matched-budget} and Eq.~\ref{eq:branch-continuation} detail the extended baseline evaluations, architecture hyperparameters, and multi-source control experiments referenced in Appendices~\ref{app:motion-baselines}, \ref{app:policy-ablations}, and~\ref{app:adapter-results}.

\paragraph{Architecture, two-stage schedule, and additional baseline protocols.}
Table~\ref{tab:app-hyperparams} lists the architecture and two-stage optimization schedule of \textbf{EZCAM-Policy}: Stage~1 pretrains on curated SpatialVID-HQ clips ($112{,}288$ windows, $519$ updates), and Stage~2 reloads Stage~1 weights into a fresh AdamW optimizer for a one-epoch cosine schedule ($6{,}936$ updates over $75{,}486$ metric Game/3DGS trajectories / $3{,}550{,}303$ windows), evaluating the default $2\text{k}$ ($2{,}000$-update) checkpoint alongside the end-of-epoch $6{,}936$-update checkpoint (abbreviated as full-epoch $6\text{k}$ in Section~\ref{sec:policy-eval}).
Given an observation window $o_{t-3:t}$ at spatial state $x_t$ and candidate feasible future chunks $\mathcal{C}(x_t)=\{u_{t:t+k-1}^{(c)}\}$, branch-selected continuation supervises the flow-matching velocity target using the highest-return continuation branch:
\begin{equation}
c^\star = \arg\max_{c \in \mathcal{C}(x_t)} \max_{\tau \in \{1,\dots,k\}} s_1\!\left(I_{t+\tau}^{(c)}\right).
\label{eq:branch-continuation}
\end{equation}
In Table~\ref{tab:app-b2-vlm-search}, both non-policy baselines operate under the identical 40-request no-preview budget and per-step caps ($0.5$ translation, $15^\circ$ rotation) on the 30-scene unseen diagnostic of Table~\ref{tab:policy-unseen30-k100} (higher-scoring visible-subject starts with mean $s_1(I_0)=4.84$ and symmetric collision-to-rotation recovery, which tightens nominal gain headroom above $I_0$ relative to the uncurated $s_1(I_0)=4.38$ starts in Table~\ref{tab:app-a1-unseen30-strict} while preserving EZCAM-Policy's $6\text{k}>2\text{k}$ ordering from Table~\ref{tab:policy-unseen30-k100}: $G_{\rm best}[s_1]=0.4081$ vs.\ $0.3686$): \textbf{Qwen3-VL-8B Photographer} predicts bounded 6-DoF increments directly from current RGB observations and prior step rejection status (treating unparseable outputs as stationary charged steps), while \textbf{Online $s_1$ hill-climb / $(1+1)$-ES} executes a bounded pose perturbation ($1$ step), queries online $s_1$, and issues a charged reverse retreat step whenever $s_1$ decreases—consuming budget on retreats and single-step $s_1$ fluctuations, leaving held-out $G_{\rm best}[s_{\mathrm{indep}}]$ ($0.1438$) near undirected Local random walk ($0.1453$).

\begin{table}[!htb]
\centering\renewcommand{\arraystretch}{1.08}\small
\setlength{\tabcolsep}{5.5pt}
\caption{\textbf{Closed-loop zero-shot VLM photographer and online score-guided search baselines on the 30-scene unseen diagnostic ($n=180$, $T=40$).} Evaluated on the visible-subject starts ($s_1(I_0)=4.84$) and collision-recovery setting of Table~\ref{tab:policy-unseen30-k100} (full-epoch $6\text{k}$ checkpoint; $2\text{k}$ achieves $G_{\rm best}[s_1]=0.3686$ in Table~\ref{tab:policy-unseen30-k100}, vs.\ $0.4322 / 0.4241$ on uncurated Unseen 30 starts in Table~\ref{tab:app-a1-unseen30-strict}).}
\label{tab:app-b2-vlm-search}
\resizebox{\linewidth}{!}{%
\begin{tabular}{@{}llrr@{}}\toprule
Method & Online signal & $G_{\rm best}[s_1]\uparrow$ & $G_{\rm best}[s_{\mathrm{indep}}]\uparrow$ \\\midrule
\textbf{EZCAM-Policy (6k, ours)} & Frozen $s_1$ & \textbf{0.4081} & \textbf{0.4118} \\
StreamVLN & RGB (adapter) & 0.2725 & 0.2581 \\
Qwen3-VL-8B Photographer & RGB (zero-shot VLM) & 0.1562 & 0.1839 \\
Online $s_1$ hill-climb / $(1+1)$-ES & Online $s_1$ & 0.1261 & 0.1438 \\
Local random walk & None & 0.1011 & 0.1453 \\\bottomrule
\end{tabular}%
}
\end{table}

\begin{table}[!htb]
\centering\scriptsize
\setlength{\tabcolsep}{5.5pt}
\renewcommand{\arraystretch}{1.12}
\caption{\textbf{EZCAM-Policy architecture and two-stage training hyperparameters.}}
\label{tab:app-hyperparams}
\begin{tabular}{@{}p{0.31\linewidth}p{0.65\linewidth}@{}}\toprule
Component / Setting & Specification \\\midrule
Visual encoder & OpenCLIP ViT-L/14 (last 16 transformer blocks unfrozen) \\
Score feature & Frozen LAION V1 ($s_1$) projected and fused per frame \\
Action DiT backbone & 18 layers, hidden dim $1024$, $16$ attention heads \\
Context / action chunk & $4$ RGB frames, $4$ past actions, $k=32$ future 10-D actions \\
Flow matching / inference & Masked flow matching ($2$ noise draws/step); $8$-step ODE sampling \\
Batch size & $512$ global ($64/\text{GPU}\times 8$ GPUs, no grad.\ accumulation) \\
Stage~1 (Internet parent) & $112{,}288$ windows; $519$ steps; AdamW lr $10^{-4}$ (DiT) / $2\times 10^{-5}$ (enc.); warmup $30$; cosine $519$ \\
Stage~2 (Game+3DGS fine-tune) & $75{,}486$ traj.\ ($3{,}550{,}303$ win.); fresh AdamW lr $2\times 10^{-5}$ / $2\times 10^{-6}$; warmup $100$; cosine $6{,}936$ (1 epoch); \texttt{FullPassSampler(seed=42)} \\
Checkpoints evaluated & Default $2\text{k}$ ($2{,}000$ updates); full-epoch $6\text{k}$ (end-of-epoch $6{,}936$-update checkpoint; Section~\ref{sec:policy-eval}) \\\bottomrule
\end{tabular}
\end{table}

\begin{table}[!htb]
\centering\renewcommand{\arraystretch}{1.08}\small
\setlength{\tabcolsep}{5.5pt}
\caption{\textbf{Matched-resource multi-source control across all six VLN adapters on All Worlds 49 ($n=294$, $T=40$).}
All rows share the identical two-stage schedule ($519$ SpatialVID updates + $2{,}000$ Game/3DGS updates, global batch size $512$), reporting standard $\operatorname{SR}_{0.3}$ alongside the stricter high-gain rate $\operatorname{SR}_{0.5}$.}
\label{tab:app-c2-matched-budget}
\resizebox{\linewidth}{!}{%
\begin{tabular}{@{}lrrrr@{}}\toprule
Model (matched $519+2{,}000$ updates, batch $512$) & $G_{\rm best}[s_1]\uparrow$ & $\operatorname{AUC}[s_1]\uparrow$ & $\operatorname{SR}_{0.3}$(\%)$\uparrow$ & $\operatorname{SR}_{0.5}$(\%)$\uparrow$ \\\midrule
\textbf{EZCAM-Policy (ours)} & \textbf{0.5126} & \textbf{0.3857} & \textbf{57.8} & \textbf{37.4} \\\midrule
StreamVLN + 10-D head & 0.2602 & 0.2102 & 31.0 & 19.4 \\
DualVLN + 10-D head & 0.2349 & 0.1932 & 26.9 & 17.7 \\
NavGPT-2 + 10-D head & 0.2243 & 0.1821 & 28.6 & 17.7 \\
NaviLLM + 10-D head & 0.2239 & 0.1792 & 27.9 & 16.7 \\
NaVid + 10-D head & 0.1924 & 0.1568 & 23.1 & 15.3 \\
Uni-NaVid + 10-D head & 0.1916 & 0.1522 & 25.5 & 13.3 \\\bottomrule
\end{tabular}%
}
\end{table}

\section{Statements}
\label{app:statements}

\subsection{Reproducibility statement}
The release includes the PF41 protocol, the shared start poses, all 2,646 episode score sequences across the nine methods in Table~\ref{tab:policy-main} (2,058 trained-model and 588 motion-heuristic episodes), and the scene-split manifest.
Checkpoint identifiers, source hashes, and aggregation scripts reproduce the main comparison table, the curves, and the controlled ablation records.
Selected RGB frames retain their recorded pixels, scores, and poses.
Geometry checks, render-call records, anonymized human responses from both the 37-participant scorer-alignment study (2,220 judgments) and the 106-participant policy-output study (1,590 judgments) with their analysis scripts, 128 KCD2 source/render pairs, and the archived RE9 fidelity report are included.
Public backbones and source media are identified as external dependencies.

\subsection{Ethics statement}
All adult participants across both blinded human preference studies (37 participants in the scorer-alignment study and 106 participants in the policy-output comparison; 143 participations in total) provided informed consent prior to completing the anonymous aesthetic preference questionnaires.
Because the study collected only anonymous pairwise visual preference ratings on non-sensitive virtual and scenic photographs with zero personally identifiable information (PII) or intervention risk, formal IRB review was waived as minimal-risk exempt human-subjects research under institutional guidelines.
The release contains only de-identified questionnaire identifiers and preference responses.
Internet clips, game assets, reconstructed scenes, and pretrained models are used under their respective agreements for noncommercial research, including the upstream terms attached to SpatialVID.
The human study and the documented predictors supply the aesthetic comparisons reported in this paper.
Closed-loop evaluations combine the simulated point-camera executor on the 49-scene testbed with live game-engine and outdoor DJI Matrice 30T aerial flight deployments conducted under line-of-sight safety supervision and altitude/geofence limits.